\documentclass[runningheads]{llncs}

\usepackage{eccv}

\usepackage{eccvabbrv}

\usepackage{graphicx}
\usepackage{booktabs}

\usepackage[accsupp]{axessibility}  % Improves PDF readability for those with disabilities.

\usepackage{hyperref}

\usepackage{orcidlink}

\usepackage{booktabs} % For better table lines
\usepackage[skip=2pt]{caption}
\usepackage{algorithm,algpseudocode,mathrsfs,subcaption}
\usepackage{subcaption}
\usepackage{bm}
\usepackage{graphicx}
\usepackage{enumitem}
\usepackage{multicol}
\usepackage{multirow}
\usepackage{xcolor}
\usepackage{colortbl}
\usepackage{xspace}
\usepackage{amsfonts}
\usepackage{bm}       % For bold Hebrew letters
\usepackage{diagbox}

\DeclareMathOperator*{\argmax}{arg\,max}
\DeclareMathOperator*{\argmin}{arg\,min}
\DeclareMathOperator*{\sgn}{sgn}
\def\bx{\mathbf{x}}

\def\btheta{\bm{\theta}}

\def\bz{\mathbf{z}}

\renewcommand{\paragraph}[1]{\par\noindent{\bf #1}}

\DeclareMathOperator{\proj}{proj}

\usepackage{wrapfig,lipsum,booktabs}
\usepackage{array}
\usepackage{changepage}

\def\eg{\emph{e.g.,}~} 
\def\ie,{\emph{i.e.,}~} \def\ie,{\emph{I.e.,}~}
\def\etal{\emph{et al}\onedot}

\begin{document}

\title{ARoFace: Alignment Robustness to Improve Low-Quality Face Recognition} 

\titlerunning{ARoFace for Low-Quality Face Recognition}

\author{Mohammad Saeed Ebrahimi Saadabadi\inst{*}\and
Sahar Rahimi Malakshan\inst{*} \and \\
Ali Dabouei\and
Nasser M. Nasrabadi}

\authorrunning{Saadabadi et al.}

\institute{West Virginia University\\ $^{*}$Equal Contribution \\
\{me00018, sr00033,Ad0046\}@mix.wvu.edu, nasser.nasrabadi@mail.wvu.edu}

\maketitle

\begin{abstract}
  Aiming to enhance Face Recognition (FR) on Low-Quality (LQ) inputs, recent studies suggest incorporating synthetic LQ samples into training. Although promising, the quality factors that are considered in these works are general rather than FR-specific, \eg, atmospheric turbulence, resolution, \etc.
  Motivated by the observation of the vulnerability of current FR models to even small Face Alignment Errors (FAE) in LQ images, we present a simple yet effective method that considers FAE as another quality factor that is tailored to FR. We seek to improve LQ FR by enhancing FR models' robustness to FAE. To this aim, we formalize the problem as a combination of differentiable spatial transformations and adversarial data augmentation in FR. We perturb the alignment of the training samples using a controllable spatial transformation and enrich the training with samples expressing FAE.
  We demonstrate the benefits of the proposed method by conducting evaluations on IJB-B, IJB-C, IJB-S (+4.3\% Rank1), and TinyFace (+2.63\%). \href{https://github.com/msed-Ebrahimi/ARoFace}{https://github.com/msed-Ebrahimi/ARoFace}
  \keywords{Unconstrained Face Recognition \and Face Alignment \and Low-Quality Input \and Adversarial Data Augmentation}
\end{abstract}

\section{Introduction}
\label{sec:intro}

Excellence in Face Recognition (FR) is attributed to large-scale training datasets \cite{guo2016ms,zhu2021webface260m}, advanced deep networks \cite{he2016deep,simonyan2014very,huang2017densely}, and angular criteria \cite{liu2017sphereface,kim2022adaface,wang2018cosface,deng2019arcface}.
However, current FR models experience performance failure in practical scenarios where images are acquired from long-range distances, \ie, Low-Quality (LQ) or unconstrained images \cite{Cheng2019Low,Kalka2018IJB,chai2023recognizability,Boutros_2023_CVPR,shi2021boosting}. 
Specifically, the reported performance \cite{shi2020towards,ali} of the state-of-the-art (SOTA) models on IJB-S \cite{Kalka2018IJB} and TinyFace \cite{Cheng2019Low}, \ie, LQ benchmarks, are about 30\% lower than LFW \cite{huang2008labeled}, \ie, High-Quality (HQ) benchmark.
This deficiency stems from the distribution gap between the training and LQ testing data, \ie, lack of a sufficient number of LQ instances during training \cite{ge2020efficient,zangeneh2020low,wang2019improved,singh2019dual,shi2021boosting,ali}.
An intuitive solution would be to construct a large-scale dataset with an adequate amount of LQ samples which is impractical due to data acquisition costs and privacy concerns \cite{Terhorst2023Qmagface,ali}.

Considering the components of a practical FR system, \ie, face detection, face alignment, and face recognition model \cite{zhao2020rdcface,wu2017recursive}, available LQ FR methods primarily have focused on how LQ inputs affect the recognition model \cite{hennings2008simultaneous,chen2018fsrnet,Li2020EnhancedBF,Wang2022RestoreFormerHB,Yang2021GANPE,ge2020efficient,wang2019improved,singh2019dual, shin2022teaching,ali}. These studies can be categorized into Facial Image Enhancement (FIE) \cite{hennings2008simultaneous,chen2018fsrnet,yu2018face,ma2020deep, Gu2022VQFRBF,Li2020EnhancedBF,Wang2022RestoreFormerHB,Yang2021GANPE}, and Common Space Mapping (CSM) \cite{ge2020efficient,zangeneh2020low,wang2019improved,singh2019dual, shin2022teaching,ali}. 
FIE estimates HQ face from LQ counterpart and then performs FR, yet faces the challenge of ill-posedness, \ie, multiple HQ outcomes exist for a single LQ face \cite{huang2017beyond, shin2022teaching,yang2014single,yang2019deep}. CSM studies \cite{ge2020efficient,zangeneh2020low,wang2019improved,singh2019dual, shin2022teaching} seek to map LQ and HQ instances to a common embedding by employing image degradations as data augmentation during training. However, recent CSM works lack practicality due to deficiency in expressing real-world LQ \cite{niemeijer2024generalization,nair2023ddpm,chai2023recognizability,nair2023ddpm}, entailing high computational resources \cite{robbins2022effect}, identity preservation challenge \cite{shin2022teaching, wang2019implicit, yang2022adversarial, li2023rethinking, antoniou2017data}, and constrained of \textit{a priori} defined target data \cite{niemeijer2024generalization,zhao2022style}.

 \begin{figure}[t]
  \centering
  \includegraphics[width=1.0\linewidth]{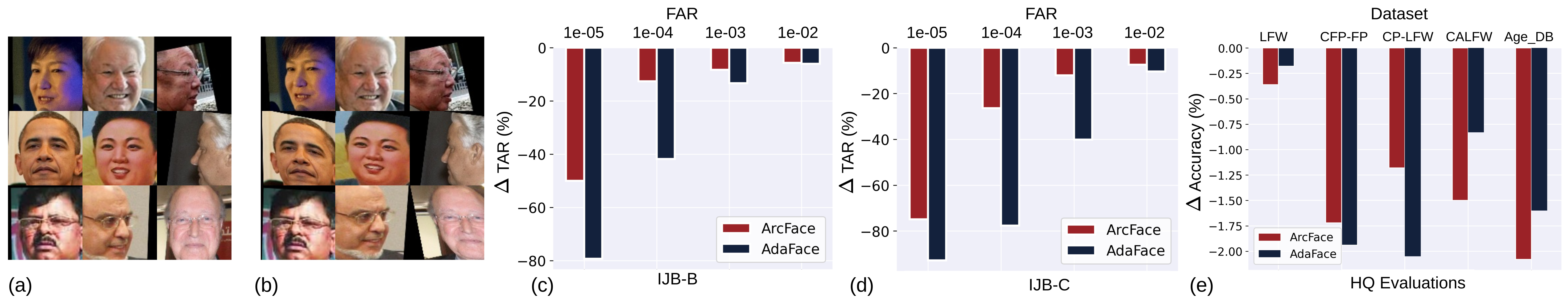}
  \caption{ 
  Visual comparison of aligned (a) and alignment-perturbed (b) samples from the IJB-B dataset. (c, d, e) 
  The performance difference between aligned inputs and those with slight FAE.
  Models exhibit robustness to FAE in HQ samples but suffer significant performance drops in LQ faces, with over 50\% reduction in $\text{TAR@FAR}=1e-5$. Results from two distinct ResNet-100 trained on MS1MV3 using ArcFace/AdaFace objective.
  } 
  \vspace{-20pt}
  \label{fig:motivation}
\end{figure}

We extend our focus beyond the face recognition model to explore the impact of LQ inputs on face detection and alignment.
It has been shown that face detector accuracy diminishes with LQ inputs, leading to Face Alignment Error (FAE) \cite{soundararajan2019machine,kim2022face,deng2020retinaface,zhang2016joint,wu2017recursive}.  
We investigate the impact of FAE on FR performance by manually adding small random spatial perturbations including scaling, rotation, and translation, to the original aligned evaluation faces, as shown in Figures \ref{fig:motivation}a, and b. The results in
As depicted in Figures \ref{fig:motivation}c, d, and e, the performance gap of a given model between alignment-perturbed and aligned samples. 
Based on these observations, we found that while SOTA FR networks are robust against FAE in HQ faces, they are susceptible to FAE in LQ samples, \eg, more than 50\% drop in True Acceptance Rate (TAR) at a False Acceptance Rate (FAR) of $1e-5$ ($\text{TAR@FAR=1e-5}$) on IJB-B and IJB-C. 

Therefore, we contend that FAE plays a crucial role in the failure of FR models on LQ faces.
Current CSM methods \cite{ge2020efficient,zangeneh2020low, shin2022teaching,ali} enrich training with synthetic LQ faces yet ignore the FAE as another degradation component. Consequently, the current FR models are extremely susceptible to the intersection of FAE and LQ input.
A potential remedy is end-to-end training of both recognition and detection networks. However, this is limited by high computational demands and the lack of large datasets with both types of labels.
A few studies tried to establish alignment-free FR by incorporating an alignment module into the recognition model \cite{jaderberg2015spatial,Zhou2018,wu2017recursive,zhong2017toward}. However, the estimated spatial transformation expresses coarse geometric information as a holistic parametric model and falls short of improving practical LQ FR \cite{kim2022face,ali,shi2021boosting}.

In an orthogonal direction, we consider FAE as another image degradation factor and aim to develop an FR model robust against FAE, dubbed Alignment Robust Face (ARoFace) recognition. 
Our proposal is inspired by the adversarial data augmentation that enhances model robustness through integrating adversarial components into the training process \cite{goodfellow2014explaining,xie2020adversarial}. 
Concretely, ARoFace employs a differentiable spatial transformation \cite{jaderberg2015spatial} to adversarially perturb the alignment of training samples and enrich the training with samples expressing FAE.

We craft samples expressing FAE by applying spatial transformations including scaling, rotation, and translation, to the benign (aligned) sample.
We employ a global transformation grid \cite{jaderberg2015spatial} that avoids face deformation to compute faithful and beneficial instances with meaningful spatial variations. 
Moreover, a randomized adversarial budget is used to improve the diversity of the crafted samples, helping to attain a variety of spatial transformations for a single input.

Compared to recent face degradation methods \cite{shi2021boosting, ali} our approach imposes negligible trainable parameters to the training pipeline, \ie, a maximum of nine additional parameters.
Moreover, the proposed method avoids the class distortion problem, \ie, changing the identity in the output face, of GAN-based face degradation by not altering the intensity of the faces and merely modifying the position of pixels \cite{shin2022teaching, wang2019implicit, yang2022adversarial, li2023rethinking, antoniou2017data}. Thus, the resulting misaligned face completely lies in the manifold of faithful faces for the FR training \cite{zhou2016view,goodfellow2014explaining,niemeijer2024generalization}. Furthermore, ARoFace circumvents the two-stage optimization and the necessity of accessing target data in previous GAN-based studies \cite{shi2021boosting,ali}.
It is worth noting that our method does not require any face or landmark detection module,
and can be readily integrated into the training pipeline of an arbitrary FR model to improve the generalization across LQ evaluations.
In summary, the contributions of the paper are as follows:
\begin{itemize}[leftmargin=*]
    \item We introduce FAE as an image degradation factor tailored for FR which has previously been ignored in LQ FR studies.
    \item We propose an optimization method that is specifically tailored to increase the FR model robustness against FAE.
    \item  We show that the proposed optimization can greatly increase the FR performance in real-world LQ evaluations such as IJB-S and TinyFace. Moreover, our framework achieves these improvements without sacrificing the performance on datasets with both HQ and LQ samples such as IJB-B and IJB-C.
    \item We empirically show that the proposed method is a plug-and-play module, providing an orthogonal improvement to SOTA FR methods.
\end{itemize}

\section{Related Works}
\subsection{Low-quality Face Recognition}
Approaches to tackle LQ FR can be categorized into: 1) Facial Image Enhancement (FIE) \cite{chen2018fsrnet,yu2018face,ma2020deep,kalarot2020component}, and 2) Common Space Mapping (CSM)  \cite{ge2020efficient,zangeneh2020low,wang2019improved,singh2019dual}.
FIE methods try to solve the inverse problem of retrieving HQ samples from their LQ counterparts and then perform FR \cite{yang2014single,yang2019deep,huang2017beyond}. Despite the success in enhancing visual quality metrics such as PSNR \cite{wang2004image}, SSIM \cite{wang2004image}, and LPIPS \cite{jo2020investigating}, they fail to increase the FR performance on real-world testing benchmarks since estimating HQ face from an LQ counterpart is an ill-posed problem, \ie, multiple HQ faces exist for single LQ input \cite{kim2019progressive,yang2019fsa,ali}. 
Conventional CSM methods try to find joint embedding for (HQ, LQ) pairs using sample-wise supervision. However, these approaches suffer from convergence issues when integrated into large-scale FR training framework \cite{wang2022efficient}.

Recently, Shi \etal proposed to employ an unlabeled LQ face dataset to establish a GAN as an image degradation module. In the follow-up work \cite{ali}, Liu \etal incorporate feedback from the FR network to the degradation so it can control the amount of image degradation to be applied to each samples. However, the requirement of \textit{a priory} selected target dataset limits their practical usability \cite{niemeijer2024generalization}.
Moreover, GANs are cumbersome to deploy during large-scale training and can introduce class distortion issues \cite{yang2022adversarial,shin2022teaching,wang2019implicit}. Also, developing a proper generative model heavily relies on prior knowledge \cite{wang2019improved,li2023rethinking,antoniou2017data,yang2022adversarial}. 

\subsection{{Adversarial Data Augmentation in Face Recognition}}
Employing adversarial images as additional training data has been extensively surfed in various deep learning applications \cite{xie2020adversarial,xie2019feature,goodfellow2014explaining,kurakin2016adversarial,madry2017towards}.
Broadly, adversarial data augmentation in FR intends to enhance the model's resilience against certain adversarial components in the inputs \cite{sharif2016accessorize,komkov2021advhat,deb2020advfaces,yin2021adv,dabouei2019fast}. Sharif et al. \cite{sharif2016accessorize} devised real-world adversarial samples by adding printed glasses to the face images, while Komkov \etal \cite{komkov2021advhat} explored adversarial hats. Deb \etal \cite{deb2020advfaces} initially leveraged GANs for synthesizing adversarial samples. Subsequently, Yin \etal \cite{yin2021adv} introduced an adversarial makeup generation framework, while Dong \etal \cite{dong2019efficient,dong2018boosting} harnessed generative models for creating adversarial attributes.
Dabouei \etal employed spatial transformation on facial landmarks to produce geometrically-perturbed adversarial faces.
Liu \etal \cite{ali} employed the domain-translation property of GAN combined with adversarial training to boost the LQ FR. However, the necessity of data collection and pre-defined target data make these methods extremely specific to a particular scenario and reduce the practicality.

\subsection{Face Alignment and Learnable Spatial Transformation}
The pioneering work of Jaderberg \etal \cite{jaderberg2015spatial} introduced a differentiable layer that performs Spatial Transformation (ST) to an input image or feature map. Owning to its intuitive transparency, the ST layer has inspired a multitude of subsequent studies in applications including, but not limited to, classification, dense correspondence matching, and FR \cite{zhou2018gridface,xu2021searching,kanazawa2016warpnet}.

Face alignment focuses on matching faces to a unified template, thereby minimizing variations in geometry that are unrelated to identity. This typically involves a 2D affine transformation to map facial landmarks to a specific 2D template \cite{Wang2018,Deng_2019_CVPR,wang2017normface,liu2017sphereface}.
Efforts to create an end-to-end FR system have led to investigations into learning flexible, non-rigid transformations for establishing an alignment-free FR system, \ie, integrating alignment into recognition module \cite{wu2017recursive, Zhou2018}. The approach by Wu \etal \cite{wu2017recursive} involved a recursive spatial transformer for complex transformation learning. Similarly, Zhou \etal \cite{Zhou2018} employed locally estimated homography transformations through a rectification network for face correction. These techniques aim to achieve alignment independence by concurrently learning alignment with the recognition process in a seamless, end-to-end manner. Taking a different approach, Xu \etal \cite{xu2021searching} explored optimizing the target template for alignment, rather than adjusting alignment parameters. However, their practical use is often hindered by computational demands and a potential decrease in the discriminatory capacity of the recognition network \cite{kim2022face}.

\begin{figure}[t]
  \centering
  \includegraphics[width=1.0\linewidth]{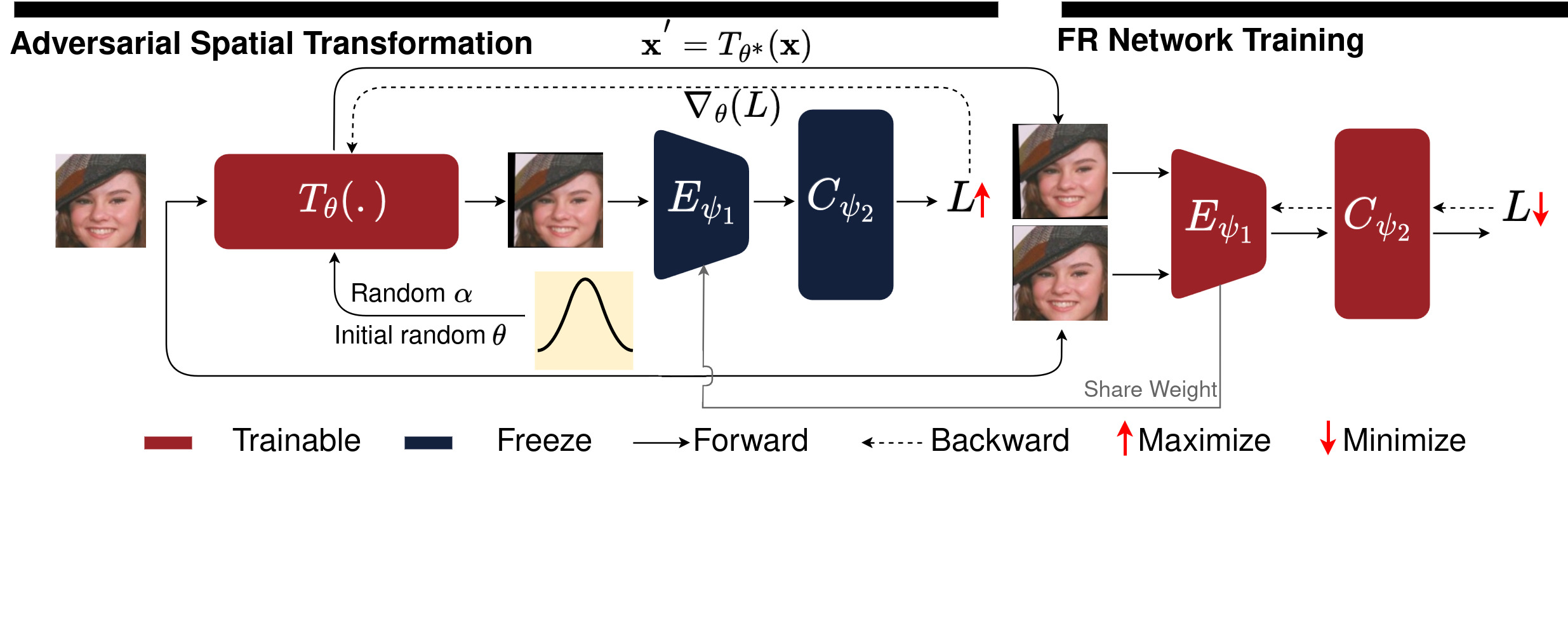}
  \vspace{-17mm}
  \caption{ Overview of proposed method. Each training iteration is composed of two steps. The adversarial spatial transformation finds $\btheta^{*} = (\varphi^{*}, \Delta {\haxis}^{*}, \Delta {\vaxis}^{*}, \lambda^{*})$ for each instance in the batch based on the feedback from the FR network to produce hard but faithful samples, \ie, maximization of $L$. Then the FR network is trained using a batch of adversarial and original samples, \ie, minimization of $L$. 
  } 
  \vspace{-15pt}
  \label{fig:main}
\end{figure} 

\section{Proposed Method}
\subsection{Notation}
In this paper, we use lowercase letters (\eg, $x$) to denote scalars, lowercase boldface (\eg, $\mathbf{x}$) to denote vectors, uppercase letters (\eg, $X$) to denote functions, and uppercase calligraphic symbols (\eg, $\mathcal{X}$) to denote sets.

Let $\mathcal{D} = \{ (\mathbf{x}_n, y_n) \}_{n=1}^{n_0}$ be the training dataset, consisting of $n_0$ aligned faces from $c$ classes. Furthermore, $F_{\psi}=C\circ E$ denotes a deep neural network with trainable parameter $\psi=[\psi_1,\psi_2]$ where $E_{\psi_1}(\cdot)\colon\mathbb{R}^q\rightarrow\mathbb{R}^d$ is the deep feature extractor that maps an input face $\mathbf{x} \in \mathbb{R}^{q}$, to a $d$-dimensional representation $\mathbf{z} \in \mathbb{R}^d$, and $C_{\psi_2}\colon\mathbb{R}^d\rightarrow \mathbb{R}^c$ is a parametric classifier, \ie, conventionally hyperspherical classifier in FR \cite{deng2019arcface,wang2018cosface,liu2017sphereface}, with parameter $\psi_2$ that maps $\bz$ to a probability distribution over $c$ classes. For brevity and convenience of presentation, all representations are $\ell_2$-normalized. 

\subsection{Preliminaries}
We propose to enhance LQ FR by increasing robustness to Face Alignment Error (FAE) since FAE in LQ images is inevitable \cite{soundararajan2019machine,kim2022face,deng2020retinaface,zhang2016joint,wu2017recursive} and current FR models are severely susceptible to FAE, as discussed in Section \ref{sec:intro}. To this end, the proposed method considers FAE as another image quality factor, \ie, tailored to FR, and leverages adversarial data augmentation combined with differentiable spatial transformation to enrich the training with samples expressing FAE. In the next two sections, we introduce adversarial data augmentation and differentiable spatial transformation, respectively, as prerequisites to the proposed method. Then, we provide a detailed description of our method.
\subsubsection{Adversarial Data Augmentation} aims to improve training without requiring additional data \cite{dabouei2019fast,peng2018jointly,madry2017towards}. \label{ada} Specifically, it seeks to craft hard training instances (adversarial samples) from the original sample $\bx$ (benign):
\begin{equation}
\label{eq:adatotal}
\theta^{*} = \argmax_{{\theta}}L(F_{\psi}; T_{\theta}(\bx), y),
\end{equation}
where $T_{\theta}(.)$ is an augmentation function with parameter $\theta$, and 
$L$ is the learning objective, \ie, empirical risk.
This optimization can be solved through $k$-steps Projected Gradient Descent (PGD) \cite{madry2017towards}:  
\begin{equation}\label{originalpgd}
\begin{aligned}
\theta_{(k+1)} \gets \proj_{\mathcal{S}} \left( \theta_{(k)} + \alpha \sgn \left(\nabla_{\theta_{(k)}} L(F_{\psi};T_{\theta_{(k)}}(\bx), y) \right)\right),
\end{aligned}
\end{equation}
where ${\sgn(.)}$ denotes the sign function, ${\alpha}$ is the PGD step size, $\mathcal{S}$ is set of allowed perturbation and $\proj_{\mathcal{S}}(\mathbf{s})$ projects $\mathbf{s}$ back into  $\mathcal{S}$, $\proj_{\mathcal{S}}(\mathbf{s})=\argmin_{\mathbf{s}^{'}\in \mathcal{S}}=||\mathbf{s}-\mathbf{s}^{'}||_2$. Typically, $\mathcal{S}$ is defined as $\ell_p$-norm ball center at $\bx$ with radius $\rho$: 
\begin{equation}\label{eq:space}
\begin{aligned}
\mathcal{S}=\{\theta\:| \: \:\:||T_{\theta}(\bx)-\bx||_{p}\leq \rho \}.
\end{aligned}
\end{equation}

\subsubsection{Spatial Transformation.}\label{STdetail}
Here, we detail the spatial transformation applied to every channel of input data, $\mathbf{x} \in \mathbb{R}^{q}$, where $q$ is $3\times h \times w$ for RGB input.
Let the row and column indices $(i, j)$ be as points $(\haxis, \vaxis) \in \mathbb{R}^2$, where the $\haxis$-axis and the $\vaxis$-axis are horizontal and vertical axis, respectively. Furthermore, $P_{\haxis}(j) = j - \frac{w-1}{2}$ and $P_{\vaxis}(i) = \frac{h-1}{2} - i$ convert zero-indexed $(i, j)$ to $\haxis, \vaxis$ coordinates.
$T_{\theta}$ is an invertible affine transformation with parameter $\theta = (\varphi, \Delta {\haxis}, \Delta {\vaxis}, \lambda)$, where $\varphi \in [0,2\pi]$ denotes the rotation angle, $\Delta{\haxis}\in\mathbb{R}$, $\Delta{\vaxis}\in\mathbb{R}$ denote horizontal and vertical shifts, respectively, and $\lambda \in \mathbb{R}$ and $\lambda > -1$ denotes the scaling factor \cite{yang2022provable}.
For each location $(P_{\haxis}(j),P_{\vaxis}(i))$ the 
coordinate that maps to this location under $T_{\theta}$ can be obtained as:
\begin{equation}\label{flowfieldrequirment}
 \small
 \begin{aligned}
  (\haxis', \vaxis') = T^{-1}_{\theta}(P_{\haxis}(j),P_{\vaxis}(i)),
\end{aligned}
\end{equation}
where $T_{\theta}^{-1}$ is the inverse transformation. $(\haxis', \vaxis')$ may not align with integer-valued pixel indices in the input image, hence, we can utilize the bi-linear interpolation kernel of Jaderberg \etal \cite{jaderberg2015spatial}:
 \begin{equation}\label{eq:interpolation}
I_{\img}(\haxis, \vaxis) = \sum_{i=0}^{h -1}\sum_{j=0}^{w -1} {\img}_{i,j}\cdot \max(0,1-|\vaxis-P_{\vaxis}(i)|)\cdot \max(0,1-|\haxis-P_{\haxis}(j)|),
\end{equation}
where $x_{i, j}$ denotes pixel value in the $i$-th row and $j$-th column of an arbitrary channel in $\bx$.
Thus the value of each pixel in the transformed image $\bx^{'}$ is:
\begin{equation}\label{eq:geometric-perturbation}
    {\img}_{i,j}^\prime = I_{\img} \big(T^{-1}_{\theta} (P_{\haxis}(j),P_{\vaxis}(i))\big),
\end{equation}
detailed derivation can be found in \cite{jaderberg2015spatial}.
With a slight abuse of notation, we term $\bx'=T_{\theta}(\bx)$ as the transformed version of $\bx$ under the transformation $T_{\theta}$. 

\subsection{Alignment Robust Face Recognition}\label{ARoface}
As observed in Figure \ref{fig:motivation}, conventional FR networks are susceptible to even mild Face Alignment Errors (FAE) in LQ faces. 
This vulnerability is a primary cause for LQ FR failure since FAE is inevitable in LQ images \cite{soundararajan2019machine,kim2022face,deng2020retinaface,zhang2016joint,wu2017recursive}. 
Previous LQ FR works \cite{shi2021boosting,ali,chen2018fsrnet,yu2018face,ge2020efficient,zangeneh2020low,wang2019improved} have ignored FAE and solely tried to make the FR model robust against general image quality factors that are not tailored to FR, \eg, resolution, atmospheric turbulence, blur, \etc.
We consider FAE as another degradation component and seek to improve LQ FR performance by making the FR model robust against FAE. To this end, the intuitive solutions are either to use unaligned faces or to apply random spatial transformations to the aligned samples. However, it has been shown that random data augmentation is ineffective in most FR evaluations \cite{ali,kim2022adaface,saadabadi2023quality}. 

\begin{algorithm}[t]
\small
\caption{ARoFace}
\label{alg:gas}
\begin{algorithmic}[1]
\Statex \textbf{Input:} Dataset $\mathcal{D}$, $\overline{f}$ for conventional five landmarks in FR, number of PGD steps $k$, distribution $\Theta$ for $\theta$, distribution $N$ for $\alpha$, and the total training iteration $t$ 
\Statex \textbf{Output:} FR network $F_{\psi}$
\State Initialize $F_{\psi}$
\For{$i=0\dots t-1 $} \Comment{\textcolor{blue}{outer optimization}}
\State Sample $\mathcal{B}$ from $\mathcal{D}$
\State Sample $\theta \sim \Theta$
\State Sample $\alpha \sim N$
\For{$j=1\dots k $}\Comment{\textcolor{blue}{inner optimization for $k$ steps}}

\State Compute $l= L(F_{\psi};T_{\theta}(\mathcal{B}_{\bx}),\mathcal{B}_{y})$
\State Update $\theta_{j+1} \leftarrow \theta_{j} +  \alpha \sgn(\nabla_{\theta_j}({l}))$
\State $\theta^{*} \leftarrow \proj_{\mathcal{S}}(\theta_{j+1})$
\EndFor

\State Compute $l_1=L(F_{\psi};T_{\theta^{*}}(\mathcal{B}_{\bx}),\mathcal{B}_{y})$ \Comment{\textcolor{blue}{{ FR objective alignment-perturbed faces}}}
\State Compute $l_2= L(F_{\psi};\mathcal{B}_{\bx},\mathcal{B}_{y})$\Comment{\textcolor{blue}{{ FR objective aligned faces}}}
\State Update $\psi \leftarrow \psi - \eta \nabla_{\psi}(l_1+l_2)$
\EndFor
\end{algorithmic}
\end{algorithm}

We aim to make the FR network robust to FAE through enriching training with samples expressing FAE. To this aim, we leverage adversarial data augmentation (detailed in Section \ref{ada}) combined with differentiable spatial transformation $T_{\theta}(.)$ \cite{jaderberg2015spatial} (composition of scaling, rotation, and translation as detailed in Section \ref{STdetail}) to craft adversarial samples expressing FAE \cite{jaderberg2015spatial}. 
Formally, we find the $\theta$ that crafts hard samples for the FR model, \ie, maximizing the FR objective function. Then, the FR network $F_{\psi}(.)$ is trained using batch formed from benign and adversarial samples, \ie, minimization of FR objective. Hence, our main optimization can be formulated as: 
\begin{equation}
\label{eq:worst}
\argmin_{\psi} {\frac{1}{|\mathcal{D}|}\sum_{(\mathbf{x},y)\in \mathcal{D}}\left[L(F_\psi; \mathbf{x}, y) + \argmax_{\theta}L(F_\psi; T_{\theta}(\mathbf{x}), y)\right]},
\end{equation}
where the maximization is solved using PDG as explained in Equation \ref{originalpgd} and $L$ is an arbitrary FR objective, \ie, ArcFace \cite{deng2019arcface}.

In crafting the adversarial sample $\bx^{'}=T_{\theta^{*}}(\bx)$, \ie, maximization in Equation \ref{eq:worst}, we do not want $\bx'$ to be indistinguishable from $\bx$, but it should lie on the manifold of valid training instances.
However, the maximization may destroy all features of the input image. For instance, $T_{\theta^{*}}$ may zero out the image, making learning infeasible. 
Typically, the $\ell_p$-norm of the difference between benign and adversarial samples is employed to define the allowed perturbation set $\mathcal{S}$, as shown in Equation \ref{eq:space}. 
This convention is based on the fact that when two images are close in a $\ell_p$-norm, they are visually similar \cite{goodfellow2014explaining}. However, its converse does not always hold, \ie, two images that are distant in a $\ell_p$-norm, can also be visually similar. For instance, large $\ell_2$-norm results from single pixel translation \cite{engstrom2019exploring}. Thus $\ell_p$-norm constrained $\mathcal{S}$ is not applicable here.

Alternatively, we utilize the per-landmark flow (displacement) vector $\mathbf{f} = (\haxis^{'}-P_{\haxis}(j),\vaxis^{'}-P_{\vaxis}(i))$ \cite{xiao2018spatially}, \ie, the vector from a position of a landmark in the $\bx^{'}$ to its corresponding position in the $\bx$, to define the $\mathcal{S}$:
\begin{equation}\label{eq:thetaspace}
\begin{aligned}
\mathcal{S}=\{\theta \:| \: \:\:\sum_{\mathbf{p} \in \mathcal{P}}||\mathbf{f}^{\theta}_{\mathbf{p}}||_{2}\leq \sum_{\mathbf{p} \in \mathcal{P}}\overline{f}_{\mathbf{p}} \},
\end{aligned}
\end{equation}
where $\mathcal{P}$ is the set of five landmarks typically employed in FR face alignment \cite{liu2017sphereface,wang2018cosface,deng2019arcface,kim2022adaface,dabouei2019fast}, and $\overline{f}_\mathbf{p}$ is the upper bound to the norm of flow vector corresponding to $\mathbf{p}$. 
Typically, FR training samples are already aligned to a predefined template, \ie, the positions of landmarks in $\bx$ are already available.
Furthermore, the final landmark location after applying the transformation can be obtained using Equation \ref{flowfieldrequirment}. Therefore, the proposed constraint does not need landmark estimation and $\mathcal{S}$ can be effectively computed using the FR alignment template, \ie, common across available large-scale datasets, and Equation \ref{flowfieldrequirment}.
Section 1 of Supplementary material provides a detailed explanation on $\mathcal{S}$.

Furthermore, to inject uncertainty to the crafted $T_{\theta}$, the PGD step size $\alpha$ is randomly sampled from the Gaussian distribution $N(\mu,\sigma^2)$:
\begin{equation}\label{finalPGD}
\resizebox{0.9\linewidth}{!}{$
\begin{aligned}
\theta_{(k+1)} \gets \proj_{\mathcal{S}} \left( \theta_{(k)} + \alpha \sgn \left(\nabla_{\theta_{(k)}} L(F_{\psi};T_{\theta_{(k)}}(\bx), y) \right)\right); \quad \text{s.t.}\:\: \alpha \sim N(\mu,\sigma^2),
\end{aligned}$}
\end{equation}
randomized $\alpha$ significantly helps the ARoFace to increase the diversity of crafted transformations. This diversity is essential to enrich the training with samples expressing the uncertainty inherent to FAE.
Following obtaining the adversarial transformation parameter using Equation \ref{finalPGD}, the $F_{\psi}$ is trained using a batch containing benign and adversarial samples, minimization of Equation \ref{eq:worst}. This framework enriches the training by spatial transformations that express FAE. Consequently, it improves the model robustness to FAE which eventually enhances the FR performance on LQ faces. Our proposal is orthogonal to the FR training objective and improves LQ FR without facing the issue of identity preservation, and reliance on a \textit{a priori} defined target dataset. The whole process is shown in Figure \ref{fig:main} and Algorithm \ref{alg:gas}.

\begin{table}[tb]
\caption{Comparison with SOTA methods on the IJB-B, IJB-C, and TinyFace when the backbone is ResNet-100. `*' indicates re-runs with official code due to missing trained models on the official repository.
  }
  \small
  \label{tab:MS1MV2}
  \centering
  \resizebox{1\linewidth}{!}{
\begin{tabular}{l|c|c|ccccc|ccccc|ccc}
\toprule 
\multirow{3}{*}{Method} & \multirow{3}{*}{Venue} & \multicolumn{1}{c|}{\multirow{3}{*}{Train Set}}   & \multicolumn{5}{c|}{{IJB-B}}                                        & \multicolumn{5}{c|}{{IJB-C}}                                        & \multicolumn{2}{c}{{TinyFace}}                                                            \\ 
                        & \multicolumn{1}{c|}{}                         &                        & \multicolumn{3}{c}{{TAR@FAR}} & \multicolumn{2}{c|}{{Identification}} & \multicolumn{3}{c}{{TAR@FAR}} & \multicolumn{2}{c|}{{Identification}} & & \multicolumn{1}{c}{} & \multicolumn{1}{c}{} \\ 
                        & \multicolumn{1}{c|}{}                         &                        & {1e-6}    & {1e-5}    & {1e-4}    & {Rank1}            & {Rank5}            & {1e-6}    & {1e-5}    & {1e-4}    & {Rank1}            & {Rank5}             & {Rank1}                & {Rank5}                         \\ \midrule 
MagFace \cite{meng2021magface}           & CVPR2021    & MS1MV2                                                      &      42.32   & 90.36   & 94.51   &     -             &        -          & {89.26}   & 93.67   & 95.81   &           -       &        -          &      -                                      &     -                                       \\
MagFace+IIC \cite{huang2024enhanced}               & ICLR2024 & MS1MV2                                                     &      -   & -   & -   &     -             &        -          & 89.38   & 93.95   & 95.89   &           -       &        -          &      -                                      &     -                                       \\
ArcFace \cite{deng2019arcface}              & CVPR2019   & MS1MV2                                                     &  38.68       &    88.50     & 94.09   &    -              &          -        &    85.66     &      92.69   & 95.74   &         -         &             -     & -                                          & -                                          \\
{ArcFace+CFSM*} \cite{ali}           & ECCV2022  & MS1MV2                                                     &    47.27     &   90.52      &    95.21     &          95.00        &     96.67             &    \textbf{90.83}     &   94.72      &   96.60      &       {96.19}           &   97.30               &  64.69                                          &               68.80                             \\
ArcFace+ARoFace         &        & MS1MV2                                                        & {\textbf{48.70}}  & \textbf{90.83}   & \textbf{95.38}   & \textbf{95.13}            & \textbf{96.71}            & \textbf{89.28}   & \textbf{94.74}   & \textbf{96.66}   & \textbf{96.20}            & \textbf{97.29}            & \textbf{67.32}                                      & \textbf{72.45}                                      \\ \midrule

ArcFace+VPL \cite{deng2021variational}             & CVPR2021    & MS1MV3                                                  &    -     &      -   & 95.56   &    -              &    -              &     -    &      -   & 96.76   &    -              &        -          &              -                              &                     -                       \\
ArcFace+SC \cite{deng2020sub}           & ECCV2020     & MS1MV3                                                    &    -    &     -   & 95.25   & -                 &    -              &  -     &  -      & 96.61   &     -             &      -            &     -                                       &                          -                  \\
ArcFace \cite{deng2019arcface}             & CVPR2019     & MS1MV3                                           & 40.27   & 92.09   & 95.47   &    95.29              &         97.01         & 90.99   & 95.31   & 96.81   &            96.61      &    97.66              & 63.81                                      & 68.05                                             \\
ArcFace+ARoFace         &    & MS1MV3                                                            & \textbf{42.31}   & \textbf{92.85}   & \textbf{95.68}   &      \textbf{95.79}            &         \textbf{97.44}         & \textbf{91.48}   & \textbf{95.69}   & \textbf{96.87}   &              \textbf{97.07}    &          \textbf{98.09}        & \textbf{67.54}                                      & \textbf{71.05}                                      \\ \midrule

ArcFace* \cite{deng2019arcface}              & CVPR2019     & WebFace4M                                                 & {44.67}   & 92.24   & 95.76   &       \textbf{96.17}           & {97.51}                 & {91.45}   & 95.43   & \textbf{97.16}   &      97.42            &          98.24        & 71.11                                      & 74.38                                      \\

ArcFace+ARoFace     & \multicolumn{1}{c|}{}    & WebFace4M                                       & \textbf{45.10}   & \textbf{92.33}   & \textbf{95.83}   &        96.13          &\textbf{97.62}                  & \textbf{91.48}   & \textbf{95.69}   & {96.87}   &        \textbf{97.51}          &        \textbf{98.35}          & \textbf{73.80}                                      & \textbf{76.53}      \\ \midrule
AdaFace \cite{kim2022adaface}             & CVPR2022      & WebFace4M                                               & \textbf{44.48}   & {92.26}   & 96.03   &       \textbf{96.26}          &97.68                  & 90.43   & 95.34   & 97.39   &       97.52          &        98.35          & 72.02                                      & 74.52      \\

AdaFace+ARoFace       & \multicolumn{1}{c|}{}   & WebFace4M                                       & {44.21}   & \textbf{93.57}   & \textbf{96.35}   &       96.23           & \textbf{97.71}                & \textbf{91.57}   & \textbf{95.98}   & \textbf{97.51}   &     \textbf{97.59}           &          \textbf{98.41}        & \textbf{73.98}                                      & \textbf{76.47}                                      \\ \midrule
AdaFace \cite{kim2022adaface}                 & CVPR2022 & WebFace12M                                                  & \textbf{47.49}   & 93.13   & \textbf{96.30}   &    \textbf{96.28}              &        97.72          & 89.47   & 95.94   & 97.54   &          97.56        &      98.38            & 72.29                                      & 74.97                                      \\

    {AdaFace+ARoFace}       & \multicolumn{1}{c|}{}   & WebFace12M                                      & {44.76}   & \textbf{93.14}   & {96.23}   &         96.23        & \textbf{97.93}                & \textbf{89.80}   & \textbf{96.20}   & \textbf{97.60}   &\textbf{97.61}                &             \textbf{98.57}     & \textbf{74.00}                                      & \textbf{76.87}     \\ 
   
    \bottomrule                          
\end{tabular}}
\vspace{-15pt}
\end{table}

\section{Experiments}
\subsection{Dataset}
We utilize the cleaned version of MS-Celeb-1M \cite{guo2016ms}, provided by \cite{deng2019arcface,deng2019lightweight} as our training dataset, \ie, MS1MV3. 
This dataset consists of almost 5M images from 90K identities. As per the conventional FR framework, all used datasets in our work are aligned and transformed to $\small{112 \times 112}$ pixels. 
We evaluate ARoFace on TinyFace \cite{cheng}, IJB-B \cite{whitelam2017iarpa}, IJB-C \cite{maze2018iarpa}, and IJB-S \cite{kalka} datasets. 

\noindent{\textbf{TinyFace}} \cite{cheng} is an FR evaluation dataset comprising 169,403 LQ face images across 5,139 identities, designed for 1:$N$ recognition. The average image size of this dataset is 20 by 16 pixels. 

\noindent\textbf{IJB-B and IJB-C.} IJB-B \cite{whitelam2017iarpa} contains approximately 21.8K images (11.8K faces and 10K non-face) and 7k videos (55K frames), representing a total of 1,845 identities.
IJB-C \cite{maze2018iarpa}, an extension of IJB-B, includes 31.3K images and 117.5K frames from 3,531 identities. IJB-B and IJB-C contain both HQ and LQ samples and have been widely used for evaluating the FR models \cite{kim2022adaface}.

\noindent\textbf{IJB-S} \cite{kalka} is recognized as one of the most challenging FR benchmarks, primarily utilizing samples from real-world surveillance videos, see Figure \ref{fig:kha}a. This dataset comprises 350 surveillance videos, totaling 30 hours, from 202 identities. Additionally, there are seven HQ photos for each subject. This dataset is characterized by three keywords:
\begin{itemize}
\item \textbf{Surveillance:} Refers to the use of surveillance video footage.
\item \textbf{Single:} Denotes the utilization of a single HQ enrollment image.
\item \textbf{Booking:} Multiple enrollment images taken from different viewpoints.
\end{itemize}

\subsection{Implimentation Details}
We adopt a modified version of ResNet-100 \cite{deng2019arcface} as our backbone. The training is done for 28 epochs using ArcFace loss, with exceptions noted where applicable. We employ SGD as the optimizer, with a cosine annealing learning rate starting from 0.1, a weight decay of 0.0001, and a momentum of 0.9.
The component of $\theta$ governing scale transformation is initialized with $N(\mu=1,\sigma^2=0.01)$, while parameters for rotation and translation are derived from $N(\mu=0,\sigma^2=0.01)$.
Furthermore, $\alpha$ is sampled from $N(\mu=0,\sigma^2=0.01)$. Section 2 of Supplementary Material provides detailed experiments on these parameters.
During training, each GPU handles a mini-batch of size 512, utilizing four Nvidia RTX 6000 GPUs. For a fair comparison, when a checkpoint for a specific method was not available, we used the official code released by the authors and the optimal hyper-parameters as recommended in their publication for reproducing their results.

\begin{table}[tb]
\caption{Comparison with SOTA methods on the IJB-S benchmark when the backbone is ResNet-100. ARoFace outperforms all the baselines by a considerable margin. 
  }
  \small
  \label{tab:IJBS}
  \centering
  \resizebox{1\linewidth}{!}{
\begin{tabular}{l|c|c|ccc|ccc|ccc}
\toprule
                                                &             &               & \multicolumn{3}{c|}{{Surveillance-to-Single}}      & \multicolumn{3}{c|}{Surveillance-to-Booking}      & \multicolumn{3}{c}{{Surveillance-to-Surveillance}}      \\
\multirow{-2}{*}{Method} & \multirow{-2}{*}{Venue}&\multirow{-2}{*}{Dataset} &  Rank1 & Rank5 & 1       & Rank1 & Rank5 & 1         & Rank1 & Rank5 & 1    \\ \midrule
ArcFace \cite{deng2019arcface}  & CVPR2019                & MS1MV2                                     & 57.35 & 64.42 & 41.85  & 57.36 & 64.95 & 41.23  & -     & -     & -        \\
PFE \cite{shi2019probabilistic}            &   ICCV2019   & MS1MV2                                      & 50.16 & 58.33 & 31.88  & 53.60 & 61.75 & 35.99  & 9.20     & 20.82     &0.84        \\
URL \cite{shi2020towards}     & ICCV2020             & MS1MV2                                     & 59.79 & 65.78 & 41.06  & 61.98 & 67.12 & 42.73  &-     & -     &-        \\
ArcFace+ARoFace        &          & MS1MV2                                  & \textbf{61.65} & \textbf{67.6} & \textbf{47.87} &  \textbf{60.66} & \textbf{67.33} & \textbf{46.34} & \textbf{18.31}     & \textbf{32.07}    & \textbf{2.23}         \\ \midrule

ArcFace \cite{deng2019arcface}     & CVPR2019              & WebFace4M                                                      &69.26 & 74.31 & 57.06  & 70.31 & 75.15 & 56.89  & 32.13     & 46.67     & 5.32         \\
ArcFace+ARoFace    &              & WebFace4M                                  & \textbf{70.96} & \textbf{75.54} & \textbf{58.67}  & \textbf{71.70} & \textbf{75.24} & \textbf{58.06} & \textbf{32.95}     & \textbf{50.30}    & \textbf{6.81}         \\ \midrule

AdaFace \cite{kim2022adaface}    & CVPR2022            & WebFace12M                             & 71.35 & 76.24 & 59.40  & 71.93 & 76.56 & 59.37  & 36.71 & 50.03 & 4.62  \\
{AdaFace+ARoFace }   &                  & WebFace12M                              & \textbf{72.28}     & \textbf{77.93}     & \textbf{61.43}        & \textbf{73.01}     & \textbf{79.11}     & \textbf{60.02}      & \textbf{40.51}     &  \textbf{50.90}   & \textbf{6.37}        \\ 
 \bottomrule  
\end{tabular}
}
\vspace{-15pt}
\end{table}

\subsection{Comparison with SOTA Methods}
Table \ref{tab:MS1MV2} compares ARoFace performance on IJB-B, IJB-C, and TinyFace against SOTA methods. These results demonstrate that ARoFace consistently sets new SOTA benchmarks across a variety of metrics and datasets. These advancements using different training datasets suggest that ARoFace's benefits are not confined to any specific training dataset. 
 When employing the WebFace4M, MS1MV3, and MS1MV2 training sets, ARoFace outperforms its competitors by 0.43\%, 2.04\%, and 1.43\% at FAR=$1e-6$ on IJB-B, respectively. Concretely, on IJB-C, ARoFace surpasses AdaFace by 1.14\% using WebFace4M and ArcFace by 0.49\% using MS1MV3 at FAR=$1e-6$.
Using MS1MV2 as the training data, ARoFace (ArcFace+ARoFace) outperforms CFSM \cite{ali} (ArcFace+CFSM) by 2.63\%, and 3.65\% improvements in Rank1, and Rank5 TinyFace identification, respectively.
These enhancements over CFSM, which prioritizes the inclusion of synthesized LQ data in the training, underscore the significance of mitigating the susceptibility to FAE over the introduction of LQ instances to the training.

\begin{table}[tb]
\caption{Comparison between baseline and baseline+ARoFace on IJB-B and IJB-C dataset on aligned and alignment-perturbed inputs.
  }
  \vspace{-3mm}
  \label{tab:robustnessimprovement}
  \centering
  \resizebox{1.0\linewidth}{!}{
   
    \begin{tabular}{lc|cccc|cccc}
    \toprule
\multicolumn{1}{l}{}                          &                         & \multicolumn{4}{c|}{IJB-B}                                                                                      & \multicolumn{4}{c}{IJB-C}                                           \\
\multicolumn{1}{l}{\multirow{-2}{*}{Method}}  & \multirow{-2}{*}{Input Alignment} & 1e-5            & 1e-4           & 1e-3           & \multicolumn{1}{c|}{1e-2}                                   & 1e-5            & 1e-4            & 1e-3           & 1e-2           \\ \midrule
\multicolumn{1}{l}{ArcFace \cite{deng2019arcface}}                   & Original                & 92.09           & 95.47          & 96.92          & \multicolumn{1}{c|}{97.84}                                  & 95.31           & 96.81           & 97.88          & 98.54          \\

\multicolumn{1}{l}{ArcFace+ARoFace}           & Original                & 92.85           & 95.68          & 96.97          & \multicolumn{1}{c|}{97.88}                                  & 95.69           & 96.87           & 97.95          & 98.56          \\ 
\rowcolor[HTML]{C0C0C0} 
\multicolumn{2}{l|}{\cellcolor[HTML]{C0C0C0}General improvement} & \textbf{+0.76} & \textbf{+0.21} & \textbf{+0.05} & \multicolumn{1}{c|}{\cellcolor[HTML]{C0C0C0}\textbf{+0.04}} & \textbf{+0.38} & \textbf{+0.06} & \textbf{+0.07} & \textbf{+0.02} \\ \hline

\multicolumn{1}{l}{ArcFace \cite{deng2019arcface}}                   & Perturbed               & 45.74           & 83.64          & 88.80          & \multicolumn{1}{c|}{92.34}                                  & 16.17           & 68.89           & 85.53          & 91.11          \\
\multicolumn{1}{l}{ArcFace+ARoFace}           & Perturbed               & 61.09           & 90.85          & 93.17          & \multicolumn{1}{c|}{94.86}                                  & 28.92           & 84.50           & 91.40          & 96.27          \\
\rowcolor[HTML]{C0C0C0} 
\multicolumn{2}{l|}{\cellcolor[HTML]{C0C0C0}Improvement to misalignment} & \textbf{+15.35} & \textbf{+7.21} & \textbf{+4.37} & \multicolumn{1}{c|}{\cellcolor[HTML]{C0C0C0}\textbf{+2.52}} & \textbf{+12.75} & \textbf{+15.61} & \textbf{+5.87} & \textbf{+5.16} \\  \midrule
 \multicolumn{1}{l}{AdaFace \cite{kim2022adaface}}                   & Original                &    90.86             &          95.84      &              97.34  & \multicolumn{1}{c|}{98.31}                                       &       95.33          &      97.09          &  98.14              &  98.89              \\
\multicolumn{1}{l}{AdaFace+ARoFace}           & Original                &    90.05             &    95.91            &    97.61            & \multicolumn{1}{c|}{98.34}                                       &    95.31            &         97.12        &          98.24      &      98.91         \\ 
\rowcolor[HTML]{C0C0C0} 
\multicolumn{2}{l|}{\cellcolor[HTML]{C0C0C0}General improvement} & \textbf{-0.81} & \textbf{+0.04} & \textbf{+0.27} & \multicolumn{1}{c|}{\cellcolor[HTML]{C0C0C0}\textbf{+0.03}} & \textbf{-0.02} & \textbf{+0.03} & \textbf{+0.1} & \textbf{+0.0.3} \\ \hline

\multicolumn{1}{l}{AdaFace \cite{kim2022adaface}}                   & Perturbed               &  15.71               &    54.88             & 84.44               & \multicolumn{1}{c|}{92.46}                                       &    2.16             &      19.12            &     57.69           &       88.14         \\
\multicolumn{1}{l}{AdaFace+ARoFace}           & Perturbed               &          48.61       &       82.44         &  89.40              & \multicolumn{1}{c|}{92.69}                                       &     20.19            &    64.52             &     85.23           &      91.61          \\
\rowcolor[HTML]{C0C0C0} 
\multicolumn{2}{l|}{\cellcolor[HTML]{C0C0C0}Improvement to misalignment} &          \textbf{+32.9}       &    \textbf{+27.56}            &     \textbf{+4.96}           &   \textbf{+0.23}                                                          &      \textbf{+18.03}           &      \textbf{+45.4}           &        \textbf{+27.54}        &    \textbf{+3.47}     \\ \bottomrule
\end{tabular}}
\vspace{-15pt}
\end{table}

\begin{figure}[t]
  \centering
 \includegraphics[width=1\textwidth]{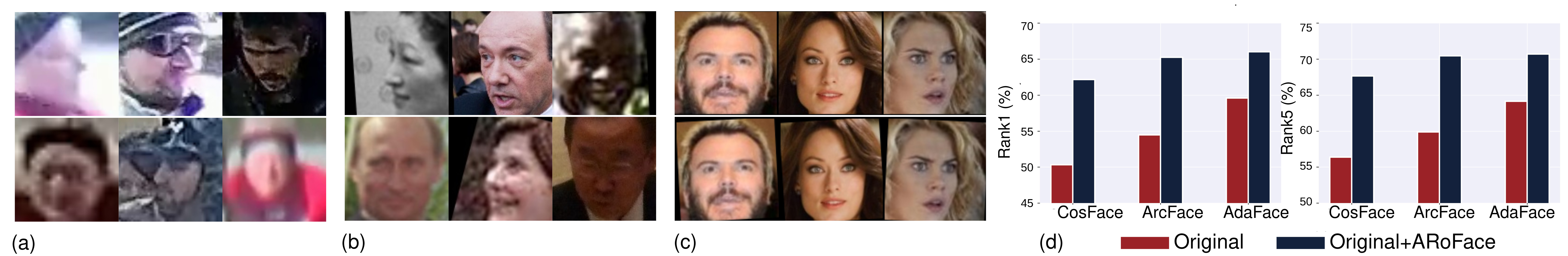}
    \caption{ (a, b) Visualizing samples from IJB-S and IJB-B datasets respectively. IJB-B consists of both HQ and LQ instances while IJB-S only consists of LQ probe instances. (c) Visualizing the benign ($\text{top}:\bx$) and their corresponding adversarial example ($\text{bottom}:T_{\theta}(\bx)$) produced by ARoFace. (d) Orthogonality of the ARoFace to different FR objective functions. In all scenarios, integrating ARoFace into training significantly improved performance on TinyFace.  
    }
    \label{fig:kha}
    \vspace{-5mm}
\end{figure}

Using WebFace4M as the training data, ARoFace boosts AdaFace performance in TinyFace evaluation by 1.82\% and 2.11\% in Rank1 and Rank5, respectively. Furthermore, using WebFace12M as the training data, ARoFace enhances AdaFace by 2.35\%, and 2.5\% in Rank1 and Rank5, respectively. These consistent enhancements demonstrate that ARoFace maintains its effectiveness as dataset size increases, \ie, from 4M to 12M.
Remarkably, ARoFace achieves top performance on TinyFace without lowering the results on IJB-B and IJB-C benchmarks, outperforming its competitors in most verification and identification metrics. IJB-B and IJB-C datasets comprise both HQ and LQ faces \cite{kim2022adaface}, as depicted in Figure \ref{fig:kha}b. Hence, consistent improvements across IJB-B, IJB-C, and TinyFace show ARoFace generalization and its ability to enhance LQ FR without compromising performance on HQ faces.

Furthermore, Table \ref{tab:IJBS} shows ARoFace's performance on the IJB-S dataset against its competitors. ARoFace establishes a new SOTA performance by significantly outperforming previous baselines. In particular, ARoFace exceeds the ArcFace baseline with the MS1MV2 training set, marking improvements of 6.02\% in Surveillance-to-Single and 5.11\% in Surveillance-to-Booking verification. This notable progress not only underscores the face alignment challenges in LQ benchmarks like IJB-S but also demonstrates our method's ability to improve FR model robustness against FAE. 
The consistent improvement across IJB-S metrics highlights the crucial role of FAE in lowering the discriminative power of existing SOTA FR models, such as ArcFace and AdaFace. These enhancements, observed with different training datasets, affirm that our method's effectiveness is independent of the training set. 

\subsection{Analysis on Robustness to Face Alignment Error}
Face alignment is an essential prerequisite for almost all available FR models. In Figure \ref{fig:motivation}, we empirically illustrate the severe vulnerability of SOTA FR networks to even mild FAE in IJB-B and IJB-C. Here, we further examine the effectiveness of ARoFace in improving model robustness against FAE. Table \ref{tab:robustnessimprovement} compares the performance of ArcFace, and AdaFace under their original training regimes and our modified training approach. Introducing ARoFace results in considerable improvement in the robustness to FAE for both methods. Specifically, the ArcFace performance on the perturbed images increases by $15.35\%$ and $12.75\%$ in TAR@FAR=$1e-5$, and $7.21\%$ and $15.61\%$ in TAR@FAR=$1e-4$ for IJB-B and IJB-C, respectively. Consistently, the Adaface performance on the perturbed images improved by $32.9\%$ and $18.03\%$ in TAR@FAR=$1e-5$, and $27.56\%$ and $45.4\%$ in TAR@FAR=$1e-4$ for IJB-B and IJB-C, respectively. 

These significant improvements showcase the efficacy of the proposed method in making the FR network more robust against FAE.
Note that the inclusion of classical augmentation techniques in the training of FR models results in decreasing model generalization \cite{kim2022adaface,saadabadi2023quality,ali}. Notably, integrating ARoFace resulted in performance improvement in aligned and alignment perturbed inputs. This consistent improvement is particularly noteworthy which emphasizes the generalizability of the proposed method and highlights that ARoFace does not sacrifice the performance on aligned samples to gain robustness against FAE.

\vspace{-3mm}
\subsection{Orthogonal Improvement to Angular Margin}
ARoFace aims to improve LQ FR by injecting misaligned samples into the training. Figure \ref{fig:kha}c shows original and augmented samples along each other. Here, we evaluate the effectiveness of ARoFace across different angular margins, depicted in Figure \ref{fig:kha}d. Specifically, we modified the training code of three SOTA methods, \ie, CosFace, ArcFace, and AdaFace, by incorporating our training policy and using the CASIA-WebFace \cite{yi2014learning} as the training data.
These results suggest the orthogonality of ARoFace to existing angular penalty losses. Thus, ARoFace can be seamlessly integrated into SOTA FR frameworks as a plug-and-play module, increasing their robustness to FAE and ultimately improving performance, particularly in LQ evaluations.

\vspace{-2mm}

\subsection{Runtime Overhead}
One significant benefit of the proposed method over \cite{ali,shi2021boosting} is its ability to bypass the need for complex image generation procedure, which requires two-step optimization, \ie, training a generator and then integrating it to the FR training. 
Our proposal crafts an affine transformation \cite{jaderberg2015spatial} to apply directly to the input image, thereby eliminating the need for two-step optimization.

\begin{figure}[t]
  \centering
 \includegraphics[width=1\textwidth]{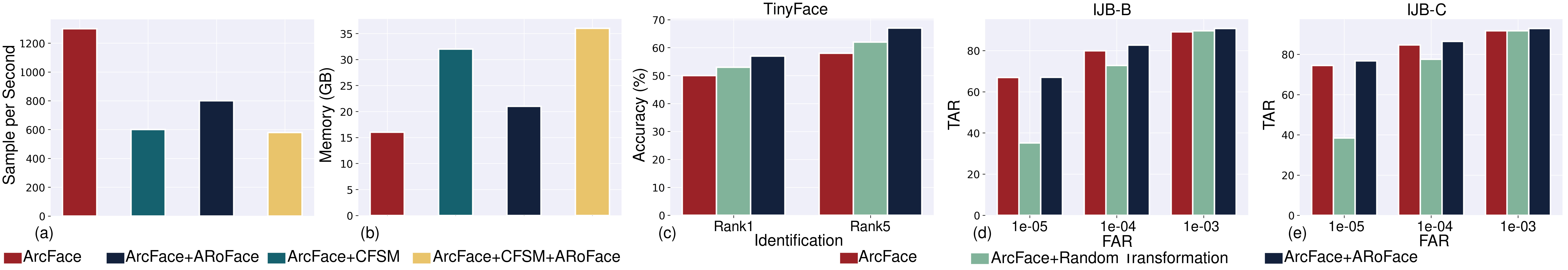}
    \caption{
    (a, b) Training speed and GPU memory consumption comparison between CFSM and ARoFace: ARoFace significantly enhances training efficiency and reduces GPU memory consumption compared to CFSM. (c, d, e) Comparing the evaluation performance between employing adversarial vs. random spatial transformation during training: Adversarial improves performance, while random fails on IJB-B and IJB-C.}
    \label{fig:trainingspeed}
    \vspace{-5mm}
\end{figure}

Moreover, ARoFace employs only four additional trainable parameters, responsible for a negligible increase in computational load. The majority of the overhead computation time and memory consumption is due to adversarial training. In Figure \ref{fig:trainingspeed}a and b, we compare our method's training speed and GPU memory usage with CFSM of Liu \etal \cite{ali}, which also aims to enhance LQ FR. By avoiding the usage of image generation models, \ie, GAN in \cite{ali}, we have increased CFSM's \cite{ali} training speed by 25\% and decreased the memory consumption by 35\%. 
These significant improvements in training speed and reduction in memory consumption, coupled with improved performance as shown in Tables \ref{tab:MS1MV2} and \ref{tab:IJBS}, emphasize the capability of our method for large-scale FR.

Furthermore, we also compare (ArcFace+CFSM+ARoFace) with the original CFSM setup (ArcFace+CFSM). CFSM already employs adversarial optimization. Hence, the integration of ARoFace into CFSM results in a negligible decrease in training speed and an increase in memory usage, as shown in Figure \ref{fig:trainingspeed}a, b. The extra computational effort from integrating ARoFace is minimal, confirming that extra computational load stems from the adversarial training which is necessary for integrating augmentation into FR training \cite{ali}.

\subsection{Ablation on Transformation Components}
\begin{wraptable}[12]{r}{0.4\textwidth}
\vspace{-10mm}
\caption{Ablation on transformation components when the training data is CASIA-WebFace, backbone is ResNet-100 and the objective function is ArcFace.
  }
  \label{tab:indbenef}
  \centering
  
 \resizebox{1\linewidth}{!}{
\begin{tabular}{ccc|c|c}
\toprule
Scale & Rotation & Translation  & Rank1 & Rank5 \\
\midrule
$\times$ & $\times$ & $\times$ &  54.47 & 59.89 \\ 
\checkmark & $\times$ & $\times$ &  59.70     &     65.17  \\
$\times$&\checkmark   & $\times$ &  57.32     &     61.93  \\
$\times$& $\times$  & \checkmark &  57.98     &     62.57  \\ 
%\midrule
\checkmark & \checkmark & $\times$ &    62.32  &    67.01   \\ 
\checkmark & \checkmark & \checkmark &  65.26    & 70.46 \\ \bottomrule      
\end{tabular}}
\end{wraptable} 
In Section \ref{ARoface}, we note that the spatial transformation employed in ARoFace comprises scaling, rotation, and translation. Here, we investigate the effect of each component on the performance of ARoFace, \ie, how much each transformation component contributes to the performance enhancement. To this end, we conduct an ablation study on TinyFace with ArcFace serving as the loss function and CASIA-WebFace \cite{yi2014learning} as the training dataset.
These experiments, as shown in Table \ref{tab:indbenef}, show that scaling is the most beneficial transformation, enhancing performance by nearly 5\%. 
This aligns with previous research \cite{ali, kim2022adaface, shi2021boosting}, highlighting scaling's role in generating LQ inputs. Translation and rotation each contribute approximately 3\% improvement, reflecting their similar influence on face alignment.

\vspace{-1.5mm}
\subsection{Random vs. Adversarial}
Liu \etal \cite{ali} showed that the performance boost is marginal without adversarial optimization, \ie, randomly generate LQ samples. Here, we investigate this effect in the proposed framework in Figure \ref{fig:trainingspeed}c, d, and e.
The results indicate that employing spatial augmentation, either random or with an adversarial signal, increases the baseline identification performance on LR inputs, \ie, TinyFace. However, random augmentation destroys the discriminative power of the FR model in HQ samples, \ie, a drastic decrease in IJB-B and IJB-C datasets. The key benefit of employing differentiable spatial transformation of ARoFace is the performance gain across datasets with different image characteristics, \ie, IJB-B, IJB-C, IJB-S, and TinyFace, and not sacrificing the performance on aligned samples to gain robustness to FAE.

\vspace{-1.5mm}
\subsection{Effect of Random Sampling of $\alpha$}
\begin{wraptable}[11]{r}{0.42\textwidth}
\vspace{-8mm}
\caption{Experiment on $\alpha$. Verification performance TAR@FAR=1e-4 is reported for IJB-B and IJB-C. Trining data is CASIA-WebFace, backbone is ResNet-100 and ArcFace is the objective function. 
  }
  \label{tab:alpha}
  \centering
  
 \resizebox{1\linewidth}{!}{

\begin{tabular}{l|cc|c|c}
\toprule
       & \multicolumn{2}{c|}{TinyFace} & \multirow{2}{*}{IJB-B} & \multirow{2}{*}{IJB-C} \\
       & Rank1         & Rank5         &                        &                        \\ \midrule
$\alpha_{fixed}$  & 62.15         & 67.48         & 58.36                  & 52.95                  \\
$\alpha_{random}$ & 65.26         & 70.46         & 58.47                  & 52.88         \\ \bottomrule       
\end{tabular}}
\end{wraptable} 
Table \ref{tab:alpha} compares the ARoFace performance with and without employing randomly sampled $\alpha$. In these experiments, when $\alpha$ is fixed, its value equals the mean of the distribution of random, \ie, $\alpha_{random}\sim N(\mu,\sigma^2)$, and  $\alpha_{fixed}=\mu$.
In TinyFace, employing $\alpha_{random}$ provides notable improvement over $\alpha_{fixed}$. This significant improvement highlights the fact that the FAE in LQ images is notably diverse and adding uncertainty through $\alpha_{random}$ helps ARoFace to increase the robustness to FAE. Furthermore, the performance on IJB-B and IJB-C is almost the same in both scenarios. We attribute this to the fact that these datasets are a mixture of LQ and HQ faces. 

\section{Conclusion}
In this paper, based on our observation on the susceptibility of current FR models to FAE in LQ faces, we proposed to consider FAE as another image degradation factor that is specifically tailored to FR. Our method employs a differentiable spatial transformation combined with adversarial data augmentation to craft samples expressing FAE and add them to the training of the FR model. This framework, allows the FR model to become exposed to FAE during the training and gain robustness against FAE.
The efficacy of the proposed method is evaluated through various experiments and evaluation across different benchmarks, including IJB-B, IJB-C, IJB-S, and TinyFace.

\bibliographystyle{splncs04}
\bibliography{main}

\begin{thebibliography}{10}
\providecommand{\url}[1]{\texttt{#1}}
\providecommand{\urlprefix}{URL }
\providecommand{\doi}[1]{https://doi.org/#1}

\bibitem{antoniou2017data}
Antoniou, A., Storkey, A., Edwards, H.: Data augmentation generative adversarial networks. arXiv preprint arXiv:1711.04340  (2017)

\bibitem{Boutros_2023_CVPR}
Boutros, F., Fang, M., Klemt, M., Fu, B., Damer, N.: Cr-fiqa: Face image quality assessment by learning sample relative classifiability. In: Proceedings of the IEEE/CVF Conference on Computer Vision and Pattern Recognition (CVPR). pp. 5836--5845 (June 2023)

\bibitem{chai2023recognizability}
Chai, J.C.L., Ng, T.S., Low, C.Y., Park, J., Teoh, A.B.J.: Recognizability embedding enhancement for very low-resolution face recognition and quality estimation. In: Proceedings of the IEEE/CVF Conference on Computer Vision and Pattern Recognition. pp. 9957--9967 (2023)

\bibitem{chen2018fsrnet}
Chen, Y., Tai, Y., Liu, X., Shen, C., Yang, J.: Fsrnet: End-to-end learning face super-resolution with facial priors. In: Proceedings of the IEEE conference on computer vision and pattern recognition. pp. 2492--2501 (2018)

\bibitem{Cheng2019Low}
Cheng, Z., Zhu, X., Gong, S.: Low-resolution face recognition. In: Computer Vision--ACCV 2018: 14th Asian Conference on Computer Vision, Perth, Australia, December 2--6, 2018, Revised Selected Papers, Part III 14. pp. 605--621. Springer (2019)

\bibitem{cheng}
Cheng, Z., Zhu, X., Gong, S.: Low-resolution face recognition. In: Computer Vision--ACCV 2018: 14th Asian Conference on Computer Vision, Perth, Australia, December 2--6, 2018, Revised Selected Papers, Part III 14. pp. 605--621. Springer (2019)

\bibitem{dabouei2019fast}
Dabouei, A., Soleymani, S., Dawson, J., Nasrabadi, N.: Fast geometrically-perturbed adversarial faces. In: 2019 IEEE Winter Conference on Applications of Computer Vision (WACV). pp. 1979--1988. IEEE (2019)

\bibitem{deb2020advfaces}
Deb, D., Zhang, J., Jain, A.K.: Advfaces: Adversarial face synthesis. In: 2020 IEEE International Joint Conference on Biometrics (IJCB). pp. 1--10. IEEE (2020)

\bibitem{deng2020sub}
Deng, J., Guo, J., Liu, T., Gong, M., Zafeiriou, S.: Sub-center arcface: Boosting face recognition by large-scale noisy web faces. In: Computer Vision--ECCV 2020: 16th European Conference, Glasgow, UK, August 23--28, 2020, Proceedings, Part XI 16. pp. 741--757. Springer (2020)

\bibitem{deng2020retinaface}
Deng, J., Guo, J., Ververas, E., Kotsia, I., Zafeiriou, S.: Retinaface: Single-shot multi-level face localisation in the wild. In: Proceedings of the IEEE/CVF conference on computer vision and pattern recognition. pp. 5203--5212 (2020)

\bibitem{deng2019arcface}
Deng, J., Guo, J., Xue, N., Zafeiriou, S.: Arcface: Additive angular margin loss for deep face recognition. In: Proceedings of the IEEE/CVF conference on computer vision and pattern recognition. pp. 4690--4699 (2019)

\bibitem{Deng_2019_CVPR}
Deng, J., Guo, J., Xue, N., Zafeiriou, S.: Arcface: Additive angular margin loss for deep face recognition. In: The IEEE Conference on Computer Vision and Pattern Recognition (CVPR) (June 2019)

\bibitem{deng2021variational}
Deng, J., Guo, J., Yang, J., Lattas, A., Zafeiriou, S.: Variational prototype learning for deep face recognition. In: Proceedings of the IEEE/CVF Conference on Computer Vision and Pattern Recognition. pp. 11906--11915 (2021)

\bibitem{deng2019lightweight}
Deng, J., Guo, J., Zhang, D., Deng, Y., Lu, X., Shi, S.: Lightweight face recognition challenge. In: Proceedings of the IEEE/CVF International Conference on Computer Vision Workshops. pp.~0--0 (2019)

\bibitem{dong2018boosting}
Dong, Y., Liao, F., Pang, T., Su, H., Zhu, J., Hu, X., Li, J.: Boosting adversarial attacks with momentum. In: Proceedings of the IEEE conference on computer vision and pattern recognition. pp. 9185--9193 (2018)

\bibitem{dong2019efficient}
Dong, Y., Su, H., Wu, B., Li, Z., Liu, W., Zhang, T., Zhu, J.: Efficient decision-based black-box adversarial attacks on face recognition. In: Proceedings of the IEEE/CVF Conference on Computer Vision and Pattern Recognition. pp. 7714--7722 (2019)

\bibitem{engstrom2019exploring}
Engstrom, L., Tran, B., Tsipras, D., Schmidt, L., Madry, A.: Exploring the landscape of spatial robustness. In: International conference on machine learning. pp. 1802--1811. PMLR (2019)

\bibitem{ge2020efficient}
Ge, S., Zhao, S., Li, C., Zhang, Y., Li, J.: Efficient low-resolution face recognition via bridge distillation. IEEE Transactions on Image Processing  \textbf{29},  6898--6908 (2020)

\bibitem{goodfellow2014explaining}
Goodfellow, I.J., Shlens, J., Szegedy, C.: Explaining and harnessing adversarial examples. arXiv preprint arXiv:1412.6572  (2014)

\bibitem{Gu2022VQFRBF}
Gu, Y., Wang, X., Xie, L., Dong, C., Li, G., Shan, Y., Cheng, M.M.: Vqfr: Blind face restoration with vector-quantized dictionary and parallel decoder. In: ECCV (2022)

\bibitem{guo2016ms}
Guo, Y., Zhang, L., Hu, Y., He, X., Gao, J.: Ms-celeb-1m: A dataset and benchmark for large-scale face recognition. In: Computer Vision--ECCV 2016: 14th European Conference, Amsterdam, The Netherlands, October 11-14, 2016, Proceedings, Part III 14. pp. 87--102. Springer (2016)

\bibitem{he2016deep}
He, K., Zhang, X., Ren, S., Sun, J.: Deep residual learning for image recognition. In: Proceedings of the IEEE conference on computer vision and pattern recognition. pp. 770--778 (2016)

\bibitem{hennings2008simultaneous}
Hennings-Yeomans, P.H., Baker, S., Kumar, B.V.: Simultaneous super-resolution and feature extraction for recognition of low-resolution faces. In: 2008 IEEE Conference on computer vision and pattern recognition. pp.~1--8. IEEE (2008)

\bibitem{huang2017densely}
Huang, G., Liu, Z., Van Der~Maaten, L., Weinberger, K.Q.: Densely connected convolutional networks. In: CVPR (2017)

\bibitem{huang2008labeled}
Huang, G.B., Mattar, M., Berg, T., Learned-Miller, E.: Labeled faces in the wild: A database forstudying face recognition in unconstrained environments. In: Workshop on faces in'Real-Life'Images: detection, alignment, and recognition (2008)

\bibitem{huang2017beyond}
Huang, R., Zhang, S., Li, T., He, R.: Beyond face rotation: Global and local perception gan for photorealistic and identity preserving frontal view synthesis. In: Proceedings of the IEEE international conference on computer vision. pp. 2439--2448 (2017)

\bibitem{huang2024enhanced}
Huang, Y., Wang, Y., Yang, L., Wang, L.: Enhanced face recognition using intra-class incoherence constraint. In: The Twelfth International Conference on Learning Representations (2024), \url{https://openreview.net/forum?id=uELjxVbrqG}

\bibitem{jaderberg2015spatial}
Jaderberg, M., Simonyan, K., Zisserman, A., et~al.: Spatial transformer networks. Advances in neural information processing systems  \textbf{28} (2015)

\bibitem{jo2020investigating}
Jo, Y., Yang, S., Kim, S.J.: Investigating loss functions for extreme super-resolution. In: Proceedings of the IEEE/CVF conference on computer vision and pattern recognition workshops. pp. 424--425 (2020)

\bibitem{kalarot2020component}
Kalarot, R., Li, T., Porikli, F.: Component attention guided face super-resolution network: Cagface. In: Proceedings of the IEEE/CVF winter conference on applications of computer vision. pp. 370--380 (2020)

\bibitem{Kalka2018IJB}
Kalka, N.D., Maze, B., Duncan, J.A., O’Connor, K., Elliott, S., Hebert, K., Bryan, J., Jain, A.K.: Ijb--s: Iarpa janus surveillance video benchmark. In: 2018 IEEE 9th international conference on biometrics theory, applications and systems (BTAS). pp.~1--9. IEEE (2018)

\bibitem{kalka}
Kalka, N.D., Maze, B., Duncan, J.A., O’Connor, K., Elliott, S., Hebert, K., Bryan, J., Jain, A.K.: Ijb--s: Iarpa janus surveillance video benchmark. In: 2018 IEEE 9th international conference on biometrics theory, applications and systems (BTAS). pp.~1--9. IEEE (2018)

\bibitem{kanazawa2016warpnet}
Kanazawa, A., Jacobs, D.W., Chandraker, M.: Warpnet: Weakly supervised matching for single-view reconstruction. In: Proceedings of the IEEE Conference on Computer Vision and Pattern Recognition. pp. 3253--3261 (2016)

\bibitem{kim2019progressive}
Kim, D., Kim, M., Kwon, G., Kim, D.S.: Progressive face super-resolution via attention to facial landmark. arXiv preprint arXiv:1908.08239  (2019)

\bibitem{kim2022face}
Kim, H.I., Yun, K., Ro, Y.M.: Face shape-guided deep feature alignment for face recognition robust to face misalignment. IEEE Transactions on Biometrics, Behavior, and Identity Science  \textbf{4}(4),  556--569 (2022)

\bibitem{kim2022adaface}
Kim, M., Jain, A.K., Liu, X.: Adaface: Quality adaptive margin for face recognition. In: Proceedings of the IEEE/CVF Conference on Computer Vision and Pattern Recognition. pp. 18750--18759 (2022)

\bibitem{komkov2021advhat}
Komkov, S., Petiushko, A.: Advhat: Real-world adversarial attack on arcface face id system. In: 2020 25th International Conference on Pattern Recognition (ICPR). pp. 819--826. IEEE (2021)

\bibitem{kurakin2016adversarial}
Kurakin, A., Goodfellow, I., Bengio, S.: Adversarial machine learning at scale. arXiv preprint arXiv:1611.01236  (2016)

\bibitem{li2023rethinking}
Li, J., Guo, Z., Li, H., Han, S., Baek, J.w., Yang, M., Yang, R., Suh, S.: Rethinking feature-based knowledge distillation for face recognition. In: Proceedings of the IEEE/CVF Conference on Computer Vision and Pattern Recognition. pp. 20156--20165 (2023)

\bibitem{Li2020EnhancedBF}
Li, X., Li, W., Ren, D., Zhang, H., Wang, M., Zuo, W.: Enhanced blind face restoration with multi-exemplar images and adaptive spatial feature fusion. CVPR  (2020)

\bibitem{ali}
Liu, F., Kim, M., Jain, A., Liu, X.: Controllable and guided face synthesis for unconstrained face recognition. In: Computer Vision--ECCV 2022: 17th European Conference, Tel Aviv, Israel, October 23--27, 2022, Proceedings, Part XII. pp. 701--719. Springer (2022)

\bibitem{liu2017sphereface}
Liu, W., Wen, Y., Yu, Z., Li, M., Raj, B., Song, L.: Sphereface: Deep hypersphere embedding for face recognition. In: Proceedings of the IEEE conference on computer vision and pattern recognition. pp. 212--220 (2017)

\bibitem{ma2020deep}
Ma, C., Jiang, Z., Rao, Y., Lu, J., Zhou, J.: Deep face super-resolution with iterative collaboration between attentive recovery and landmark estimation. In: Proceedings of the IEEE/CVF conference on computer vision and pattern recognition. pp. 5569--5578 (2020)

\bibitem{madry2017towards}
Madry, A., Makelov, A., Schmidt, L., Tsipras, D., Vladu, A.: Towards deep learning models resistant to adversarial attacks. arXiv preprint arXiv:1706.06083  (2017)

\bibitem{maze2018iarpa}
Maze, B., Adams, J., Duncan, J.A., Kalka, N., Miller, T., Otto, C., Jain, A.K., Niggel, W.T., Anderson, J., Cheney, J., et~al.: Iarpa janus benchmark-c: Face dataset and protocol. In: 2018 international conference on biometrics (ICB). pp. 158--165. IEEE (2018)

\bibitem{meng2021magface}
Meng, Q., Zhao, S., Huang, Z., Zhou, F.: Magface: A universal representation for face recognition and quality assessment. In: Proceedings of the IEEE/CVF Conference on Computer Vision and Pattern Recognition. pp. 14225--14234 (2021)

\bibitem{nair2023ddpm}
Nair, N.G., Mei, K., Patel, V.M.: At-ddpm: Restoring faces degraded by atmospheric turbulence using denoising diffusion probabilistic models. In: Proceedings of the IEEE/CVF Winter Conference on Applications of Computer Vision. pp. 3434--3443 (2023)

\bibitem{niemeijer2024generalization}
Niemeijer, J., Schwonberg, M., Term{\"o}hlen, J.A., Schmidt, N.M., Fingscheidt, T.: Generalization by adaptation: Diffusion-based domain extension for domain-generalized semantic segmentation. In: Proceedings of the IEEE/CVF Winter Conference on Applications of Computer Vision. pp. 2830--2840 (2024)

\bibitem{peng2018jointly}
Peng, X., Tang, Z., Yang, F., Feris, R.S., Metaxas, D.: Jointly optimize data augmentation and network training: Adversarial data augmentation in human pose estimation. In: Proceedings of the IEEE conference on computer vision and pattern recognition. pp. 2226--2234 (2018)

\bibitem{robbins2022effect}
Robbins, W., Boult, T.E.: On the effect of atmospheric turbulence in the feature space of deep face recognition. In: Proceedings of the IEEE/CVF Conference on Computer Vision and Pattern Recognition. pp. 1618--1626 (2022)

\bibitem{saadabadi2023quality}
Saadabadi, M.S.E., Malakshan, S.R., Zafari, A., Mostofa, M., Nasrabadi, N.M.: A quality aware sample-to-sample comparison for face recognition. In: Proceedings of the IEEE/CVF Winter Conference on Applications of Computer Vision. pp. 6129--6138 (2023)

\bibitem{sharif2016accessorize}
Sharif, M., Bhagavatula, S., Bauer, L., Reiter, M.K.: Accessorize to a crime: Real and stealthy attacks on state-of-the-art face recognition. In: Proceedings of the 2016 acm sigsac conference on computer and communications security. pp. 1528--1540 (2016)

\bibitem{shi2019probabilistic}
Shi, Y., Jain, A.K.: Probabilistic face embeddings. In: Proceedings of the IEEE/CVF International Conference on Computer Vision. pp. 6902--6911 (2019)

\bibitem{shi2021boosting}
Shi, Y., Jain, A.K.: Boosting unconstrained face recognition with auxiliary unlabeled data. In: Proceedings of the IEEE/CVF Conference on Computer Vision and Pattern Recognition. pp. 2795--2804 (2021)

\bibitem{shi2020towards}
Shi, Y., Yu, X., Sohn, K., Chandraker, M., Jain, A.K.: Towards universal representation learning for deep face recognition. In: Proceedings of the IEEE/CVF Conference on Computer Vision and Pattern Recognition. pp. 6817--6826 (2020)

\bibitem{shin2022teaching}
Shin, S., Lee, J., Lee, J., Yu, Y., Lee, K.: Teaching where to look: Attention similarity knowledge distillation for low resolution face recognition. In: Computer Vision--ECCV 2022: 17th European Conference, Tel Aviv, Israel, October 23--27, 2022, Proceedings, Part XII. pp. 631--647. Springer Nature Switzerland Cham (2022)

\bibitem{simonyan2014very}
Simonyan, K., Zisserman, A.: Very deep convolutional networks for large-scale image recognition. arXiv preprint arXiv:1409.1556  (2014)

\bibitem{singh2019dual}
Singh, M., Nagpal, S., Singh, R., Vatsa, M.: Dual directed capsule network for very low resolution image recognition. In: Proceedings of the IEEE/CVF International Conference on Computer Vision. pp. 340--349 (2019)

\bibitem{soundararajan2019machine}
Soundararajan, R., Biswas, S.: Machine vision quality assessment for robust face detection. Signal Processing: Image Communication  \textbf{72},  92--104 (2019)

\bibitem{Terhorst2023Qmagface}
Terh{\"o}rst, P., Ihlefeld, M., Huber, M., Damer, N., Kirchbuchner, F., Raja, K., Kuijper, A.: Qmagface: Simple and accurate quality-aware face recognition. In: Proceedings of the IEEE/CVF Winter Conference on Applications of Computer Vision. pp. 3484--3494 (2023)

\bibitem{wang2017normface}
Wang, F., Xiang, X., Cheng, J., Yuille, A.L.: Normface: L2 hypersphere embedding for face verification. In: Proceedings of the 25th ACM international conference on Multimedia. pp. 1041--1049 (2017)

\bibitem{wang2018cosface}
Wang, H., Wang, Y., Zhou, Z., Ji, X., Gong, D., Zhou, J., Li, Z., Liu, W.: {CosFace}: Large margin cosine loss for deep face recognition. In: CVPR (2018)

\bibitem{Wang2018}
Wang, H., Wang, Y., Zhou, Z., Ji, X., Gong, D., Zhou, J., Li, Z., Liu, W.: {CosFace: Large Margin Cosine Loss for Deep Face Recognition}. Proceedings of the IEEE Computer Society Conference on Computer Vision and Pattern Recognition pp. 5265--5274 (2018)

\bibitem{wang2022efficient}
Wang, K., Wang, S., Zhang, P., Zhou, Z., Zhu, Z., Wang, X., Peng, X., Sun, B., Li, H., You, Y.: An efficient training approach for very large scale face recognition. In: Proceedings of the IEEE/CVF Conference on Computer Vision and Pattern Recognition. pp. 4083--4092 (2022)

\bibitem{wang2019improved}
Wang, M., Liu, R., Hajime, N., Narishige, A., Uchida, H., Matsunami, T.: Improved knowledge distillation for training fast low resolution face recognition model. In: Proceedings of the IEEE/CVF International Conference on Computer Vision Workshops. pp.~0--0 (2019)

\bibitem{wang2019implicit}
Wang, Y., Pan, X., Song, S., Zhang, H., Huang, G., Wu, C.: Implicit semantic data augmentation for deep networks. Advances in Neural Information Processing Systems  \textbf{32} (2019)

\bibitem{wang2004image}
Wang, Z., Bovik, A.C., Sheikh, H.R., Simoncelli, E.P.: Image quality assessment: from error visibility to structural similarity. IEEE transactions on image processing  \textbf{13}(4),  600--612 (2004)

\bibitem{Wang2022RestoreFormerHB}
Wang, Z., Zhang, J., Chen, R., Wang, W., Luo, P.: Restoreformer: High-quality blind face restoration from undegraded key-value pairs. CVPR  (2022)

\bibitem{whitelam2017iarpa}
Whitelam, C., Taborsky, E., Blanton, A., Maze, B., Adams, J., Miller, T., Kalka, N., Jain, A.K., Duncan, J.A., Allen, K., et~al.: Iarpa janus benchmark-b face dataset. In: proceedings of the IEEE conference on computer vision and pattern recognition workshops. pp. 90--98 (2017)

\bibitem{wu2017recursive}
Wu, W., Kan, M., Liu, X., Yang, Y., Shan, S., Chen, X.: Recursive spatial transformer (rest) for alignment-free face recognition. In: Proceedings of the IEEE International Conference on Computer Vision. pp. 3772--3780 (2017)

\bibitem{xiao2018spatially}
Xiao, C., Zhu, J.Y., Li, B., He, W., Liu, M., Song, D.: Spatially transformed adversarial examples. arXiv preprint arXiv:1801.02612  (2018)

\bibitem{xie2020adversarial}
Xie, C., Tan, M., Gong, B., Wang, J., Yuille, A.L., Le, Q.V.: Adversarial examples improve image recognition. In: Proceedings of the IEEE/CVF conference on computer vision and pattern recognition. pp. 819--828 (2020)

\bibitem{xie2019feature}
Xie, C., Wu, Y., Maaten, L.v.d., Yuille, A.L., He, K.: Feature denoising for improving adversarial robustness. In: Proceedings of the IEEE/CVF conference on computer vision and pattern recognition. pp. 501--509 (2019)

\bibitem{xu2021searching}
Xu, X., Meng, Q., Qin, Y., Guo, J., Zhao, C., Zhou, F., Lei, Z.: Searching for alignment in face recognition. In: Proceedings of the AAAI Conference on Artificial Intelligence. vol.~35, pp. 3065--3073 (2021)

\bibitem{yang2014single}
Yang, C.Y., Ma, C., Yang, M.H.: Single-image super-resolution: A benchmark. In: Computer Vision--ECCV 2014: 13th European Conference, Zurich, Switzerland, September 6-12, 2014, Proceedings, Part IV 13. pp. 372--386. Springer (2014)

\bibitem{yang2022adversarial}
Yang, K., Sun, Y., Su, J., He, F., Tian, X., Huang, F., Zhou, T., Tao, D.: Adversarial auto-augment with label preservation: A representation learning principle guided approach. Advances in Neural Information Processing Systems  \textbf{35},  22035--22048 (2022)

\bibitem{yang2022provable}
Yang, R., Laurel, J., Misailovic, S., Singh, G.: Provable defense against geometric transformations. In: The Eleventh International Conference on Learning Representations (2022)

\bibitem{Yang2021GANPE}
Yang, T., Ren, P., Xie, X., Zhang, L.: Gan prior embedded network for blind face restoration in the wild. CVPR  (2021)

\bibitem{yang2019fsa}
Yang, T.Y., Chen, Y.T., Lin, Y.Y., Chuang, Y.Y.: Fsa-net: Learning fine-grained structure aggregation for head pose estimation from a single image. In: Proceedings of the IEEE/CVF conference on computer vision and pattern recognition. pp. 1087--1096 (2019)

\bibitem{yang2019deep}
Yang, W., Zhang, X., Tian, Y., Wang, W., Xue, J.H., Liao, Q.: Deep learning for single image super-resolution: A brief review. IEEE Transactions on Multimedia  \textbf{21}(12),  3106--3121 (2019)

\bibitem{yi2014learning}
Yi, D., Lei, Z., Liao, S., Li, S.Z.: Learning face representation from scratch. arXiv preprint arXiv:1411.7923  (2014)

\bibitem{yin2021adv}
Yin, B., Wang, W., Yao, T., Guo, J., Kong, Z., Ding, S., Li, J., Liu, C.: Adv-makeup: A new imperceptible and transferable attack on face recognition. arXiv preprint arXiv:2105.03162  (2021)

\bibitem{yu2018face}
Yu, X., Fernando, B., Ghanem, B., Porikli, F., Hartley, R.: Face super-resolution guided by facial component heatmaps. In: Proceedings of the European conference on computer vision (ECCV). pp. 217--233 (2018)

\bibitem{zangeneh2020low}
Zangeneh, E., Rahmati, M., Mohsenzadeh, Y.: Low resolution face recognition using a two-branch deep convolutional neural network architecture. Expert Systems with Applications  \textbf{139},  112854 (2020)

\bibitem{zhang2016joint}
Zhang, K., Zhang, Z., Li, Z., Qiao, Y.: Joint face detection and alignment using multitask cascaded convolutional networks. IEEE signal processing letters  \textbf{23}(10),  1499--1503 (2016)

\bibitem{zhao2020rdcface}
Zhao, H., Ying, X., Shi, Y., Tong, X., Wen, J., Zha, H.: Rdcface: radial distortion correction for face recognition. In: Proceedings of the IEEE/CVF Conference on Computer Vision and Pattern Recognition. pp. 7721--7730 (2020)

\bibitem{zhao2022style}
Zhao, Y., Zhong, Z., Zhao, N., Sebe, N., Lee, G.H.: Style-hallucinated dual consistency learning for domain generalized semantic segmentation. In: European Conference on Computer Vision. pp. 535--552. Springer Nature Switzerland Cham (2022)

\bibitem{zhong2017toward}
Zhong, Y., Chen, J., Huang, B.: Toward end-to-end face recognition through alignment learning. IEEE signal processing letters  \textbf{24}(8),  1213--1217 (2017)

\bibitem{Zhou2018}
Zhou, E., Cao, Z., Sun, J.: Gridface: Face rectification via learning local homography transformations. In: Proceedings of the European Conference on Computer Vision (ECCV). pp. 3--19 (2018)

\bibitem{zhou2018gridface}
Zhou, E., Cao, Z., Sun, J.: Gridface: Face rectification via learning local homography transformations. In: Proceedings of the European conference on computer vision (ECCV). pp. 3--19 (2018)

\bibitem{zhou2016view}
Zhou, T., Tulsiani, S., Sun, W., Malik, J., Efros, A.A.: View synthesis by appearance flow. In: Computer Vision--ECCV 2016: 14th European Conference, Amsterdam, The Netherlands, October 11--14, 2016, Proceedings, Part IV 14. pp. 286--301. Springer (2016)

\bibitem{zhu2021webface260m}
Zhu, Z., Huang, G., Deng, J., Ye, Y., Huang, J., Chen, X., Zhu, J., Yang, T., Lu, J., Du, D., et~al.: {WebFace260M}: A benchmark unveiling the power of million-scale deep face recognition. In: CVPR (2021)

\end{thebibliography}

\newpage
\section*{Supplementary Materials}
\setcounter{section}{0}
\section{Defining Allowed Perturbation Set $\mathcal{S}$}

Conventionally face detection and alignment are performed before training the FR module, ensuring the consistency of landmarks' positions across all training instances according to the alignment template. Therefore, the initial positions of landmarks $(u_p,v_p)$ match that of the alignment template. Additionally, one can determine the final positions of landmarks after applying the invertible affine transformation $T_{\theta}$, as described in Equation 4 of Manuscript. Consequently, the displacement vector's upper bound $\overline{\mathbf{f}}$ is calculable by considering the transformation components' upper bounds, namely the maximum permissible rotation, translation, and scaling:
\begin{align}
\overline{\mathbf{f}}_p = (u_p^{'}-u_p,v_p^{'}-v_p); \quad  (\haxis', \vaxis') = T^{-1}_{\overline{\theta}}(P_{\haxis}(j),P_{\vaxis}(i)),
\end{align}%$}
where $\overline{\theta}$ denotes the upper bound for the transformation parameters, \textit{i.e}, maximum rotation, scaling and translation. 
In the same way, the displacement vector is calculated during the training for defining $\mathcal{S}$:
\begin{align}
{\mathbf{f}}_p^{\theta} = (u_p^{'}-u_p,v_p^{'}-v_p); \quad  (\haxis', \vaxis') = T^{-1}_{{\theta}}(P_{\haxis}(j),P_{\vaxis}(i)).
\end{align}%$}
Finally, $\overline{f}_p = ||\overline{\mathbf{f}}_p||_2$ and we define the allowed transformation using Equation 8 of Manuscript without requiring landmark estimation.

\vspace{-4.5mm}

\subsection{Experiment on Upper Bound of Individual Transformations}
\begin{wraptable}[11]{r}{0.6\textwidth}
\vspace{-7mm}
\caption{Experiments on the upper bound of affine transformations. TinyFace performance is the Rank1 identification. Verification performance TAR@FAR=1e-4 is reported for IJB-B and IJB-C. Training data is CASIA-WebFace, backbone is ResNet-100 and ArcFace is the objective function. 
  }
  \label{tab:upperbound}
  \centering
  
 \resizebox{1\linewidth}{!}{
\begin{tabular}{l|ccc|ccc|ccc}
\toprule
         & \multicolumn{3}{c|}{Scaling} & \multicolumn{3}{c|}{Rotation} & \multicolumn{3}{c}{Translation} \\
         & 0.01     & 0.1     & 0.5     & 0.01     & 0.1      & 0.2     & 0.01      & 0.1      & 0.2      \\ \midrule
TinyFace & 61.37    & 62.11   & 62.24   & 50.32    & 53.76    & 57.26   & 54.85     & 57.31    & 56.18    \\
IJB-B    & 58.40   & 35.17   & 16.92   & 55.68    & 37.03    & 1.76    & 46.56     & 9.49     & 2.60     \\
IJB-C    & 52.90    & 48.84   & 41.74   & 41.49    & 16.20    & 5.71    & 52.84     & 36.94    & 6.36     \\ \bottomrule
\end{tabular}}
\end{wraptable} 
With the predefined face alignment template, the allowed perturbation set $\mathcal{S}$ is directly linked to $\overline{\theta}$, which defines the maximum allowed values for individual transformations. Here, in Table \ref{tab:upperbound}, we conduct experiments on the upper bounds of the components of transformation $T_{\theta}$, \ie, scaling, translation, and rotation. The results indicate consistent improvement in TinyFace performance with increased upper bounds across different components. Furthermore, scaling has a more significant impact on TinyFace performance, aligning with previous research on the effects of augmentation in FR performance \cite{ali, kim2022adaface, shi2021boosting}. Moreover, the results suggest that the performance on datasets containing both HQ and LQ instances is susceptible to the high values of transformation upper bounds, leading to a significant reduction in performance. We attribute this to the nature of the images in these datasets. Since both HQ and LQ samples are present in these benchmarks, excessive augmentation reduces the discriminative power of the FR model on HQ images, thereby diminishing performance in these evaluations.

 \begin{figure}[tb]
  \centering
  \includegraphics[width=1.\linewidth]{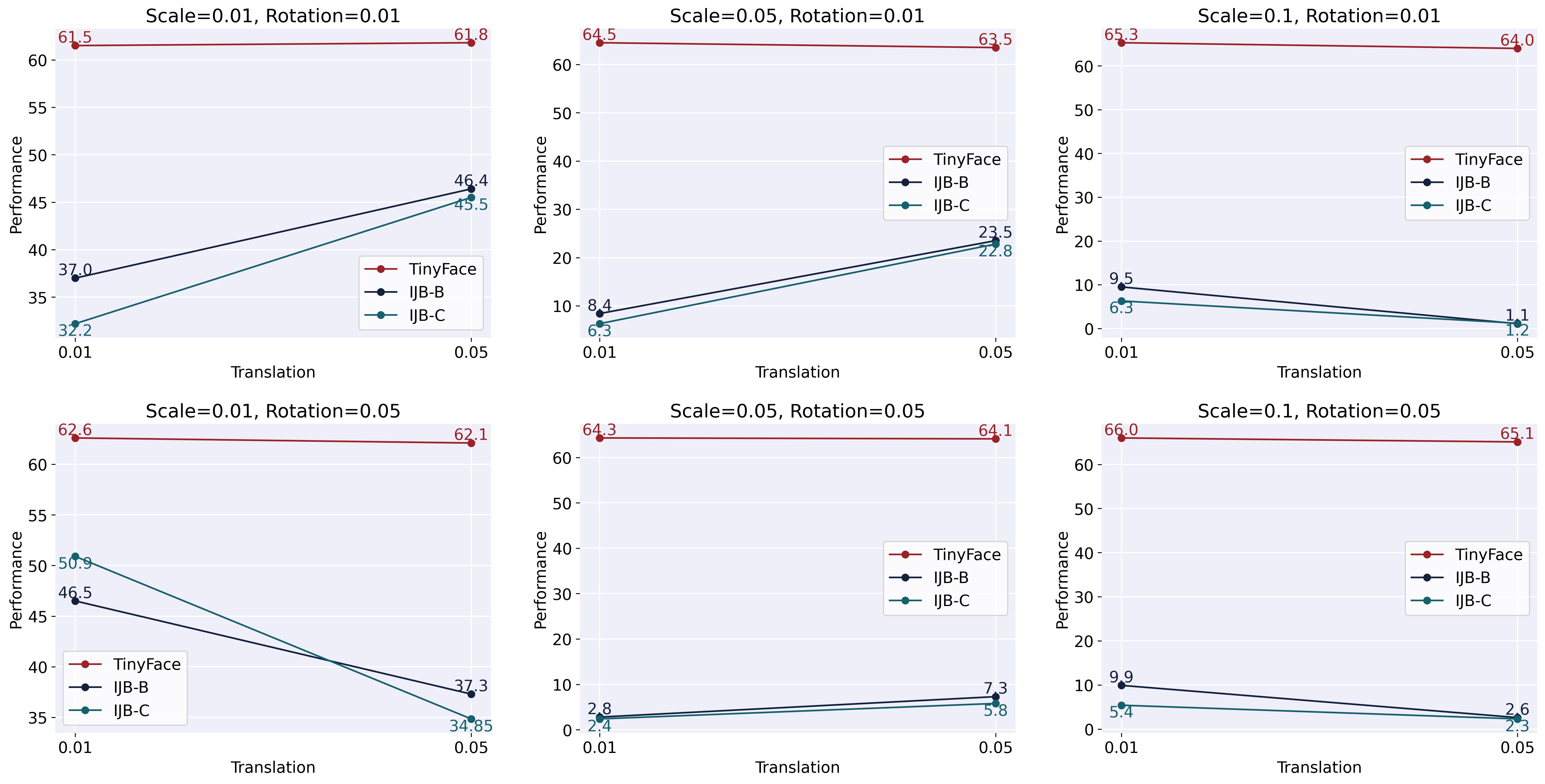}
  \caption{ 
  Experiments on different perturbation budgets. Tinyface performance is the Rank1 identification. IJB-B and IJB-C performance is the TAR@FAR=$1e-5$.
  } 
  \vspace{-15pt}
  \label{fig:motivationsup}
\end{figure}

\section{Experiments on Perturbation Budgets}
We conduct experiments to evaluate the impact of perturbation budgets $\alpha$, corresponding to the components of spatial transformation, \textit{i.e}, scale, rotation, and translation, as shown in Figure \ref{fig:motivationsup}. 
We use CASIA-WebFace as the training dataset, and ArcFace serves as the objective. The results demonstrate a positive correlation between the scaling budget and TinyFace performance; specifically, a larger scaling budget significantly enhances Rank1 TinyFace identification. This outcome is anticipated, as larger scaling perturbations produce smaller faces, thereby improving performance on LQ samples.

However, a consistent increase in the scaling budget leads to a significant decrease in verification performance on the IJB-B and IJB-C datasets. We expected these results since IJB-B and IJB-C consist of both HQ and LQ images and severe scaling results in reducing the discriminative power on the HQ images \cite{kim2022adaface,ali,saadabadi2023quality}. 
Aiming for a generalizable FR module, we have selected a scaling budget of 0.01. Furthermore, a rotation budget of 0.01 results in superior performance across IJB-B and IJB-C evaluation, indicating potential misalignment in HQ samples. 
Additionally, the translation budget of 0.01 results in balanced performance across TinyFace and other datasets. CNNs are somewhat translation-invariant, partly due to the nature of pooling operations. Thus, this minor perturbation aids in enhancing the model's robustness to misalignment. It is important to note that extreme translation might zero out essential parts of the face, rendering learning infeasible, unlike rotation, which does not remove any part of the image.
\newpage
\section{Experiments on PGD Steps $k$}
\begin{wraptable}[7]{r}{0.3\textwidth}
\vspace{-10mm}
\caption{Experiments on the number of PGD steps.
  }
  \label{tab:k}
  \centering
  
 \resizebox{1\linewidth}{!}{
\begin{tabular}{l|c|c|c}
\toprule
$k$ & TinyFace & IJB-B & IJB-C \\ \midrule
1 & 65.26    & 58.47 & 52.88 \\
2 & 65.28    & 58.54 & 52.68 \\
3 & 65.35    & 58.50 & 52.48
\\ \bottomrule\end{tabular}}
\end{wraptable} 
In Table \ref{tab:k} we study the impact of $k$ on the performance across TinyFace, IJB-B, and IJB-C. We experiment using $k \in \{1,2,3\}$. As Table \ref{tab:k} shows, the performance gain with the increase in the PGD steps is marginal. However, the computational overhead increases drastically with the increase in the number of PGD steps (detailed in Section 4.6 of Manuscript). Therefore, we employ $k=1$ across our experiments. The marginal reduction of the performance on IJB-B and IJB-C is in line with observations of \cite{ali,kim2022adaface,saadabadi2023quality} that these datasets contain both HQ and LQ instances and too much augmentations of the input, results in performance reduction in them.

\end{document}